\pdfoutput=1

\documentclass{article} 
\usepackage{iclr2024_conference,times}

\usepackage{amsmath,amsfonts,bm}

\def\eqref#1{equation~\ref{#1}}

\def\1{\bm{1}}

\DeclareMathAlphabet{\mathsfit}{\encodingdefault}{\sfdefault}{m}{sl}
\SetMathAlphabet{\mathsfit}{bold}{\encodingdefault}{\sfdefault}{bx}{n}

\usepackage{url}

\usepackage{bm}
\usepackage{times}
\usepackage{epsfig}
\usepackage{amsmath}
\usepackage{subfigure}
\usepackage{wrapfig}

\usepackage{setspace}
\usepackage{graphicx}
\usepackage{booktabs}
\usepackage{amssymb}
\usepackage{makecell}
\usepackage{verbatim}
\usepackage[ruled, lined, boxed]{algorithm2e}
\usepackage{multirow}
\usepackage{enumitem}
\usepackage{microtype}
\usepackage[breaklinks=true,bookmarks=false]{hyperref}
\usepackage[capitalize]{cleveref}
\usepackage{bbding}

\usepackage{algorithmic}

\usepackage[utf8]{inputenc} 
\usepackage[T1]{fontenc}    
\usepackage{hyperref}       
\usepackage{url}            
\usepackage{booktabs}       
\usepackage{amsfonts}       
\usepackage{nicefrac}       
\usepackage{microtype}      
\usepackage{xcolor}         

\def\eg{\emph{e.g.}} 
\def\ie{\emph{i.e.}} 

\title{Few-shot Hybrid Domain Adaptation of Image Generators}

\author{
Hengjia Li$^{1}$, ~Yang Liu$^{1}$, ~Linxuan Xia$^{1}$, ~Yuqi Lin$^{1}$, ~Tu Zheng$^{2}$, ~Zheng Yang$^{2}$, ~Wenxiao Wang$^{1}$, \And~Xiaohui Zhong$^{3}$, ~Xiaobo Ren$^{3}$, ~Xiaofei He$^{1,2}$
\\[.253cm]
{
\textsuperscript{1} {Zhejiang University} \;
\textsuperscript{2} {Fabu Inc.} \;
\textsuperscript{3} {Ningbo Port} \;
}
}

\iclrfinalcopy 
\begin{document}

\maketitle

\begin{abstract}
Can a pre-trained generator be adapted to the hybrid of multiple target domains and generate images with integrated attributes of them? In this work, we introduce a new task -- Few-shot \textit{Hybrid Domain Adaptation} (HDA). Given a source generator and several target domains, HDA aims to acquire an adapted generator that preserves the integrated attributes of all target domains, without overriding the source domain's characteristics. 
Compared with \textit{Domain Adaptation} (DA), HDA offers greater flexibility and versatility to adapt generators to more composite and expansive domains. Simultaneously, HDA also presents more challenges than DA as we have access only to images from individual target domains and lack authentic images from the hybrid domain.
To address this issue, we introduce a discriminator-free framework that directly encodes different domains' images into well-separable subspaces. To achieve HDA, we propose a novel directional subspace loss comprised of a distance loss and a direction loss. Concretely, the distance loss blends the attributes of all target domains by reducing the distances from generated images to all target subspaces. The direction loss preserves the characteristics from the source domain by guiding the adaptation along the perpendicular to subspaces. 
Experiments show that our method can obtain numerous domain-specific attributes in a single adapted generator, which surpasses the baseline methods in semantic similarity, image fidelity, and cross-domain consistency. The code will be available \href{https://github.com/EchoPluto/FHDA}{here}.
\end{abstract}

\section{Introduction}
\label{intro}
\begin{figure}
    \centering
    \includegraphics[width=0.90\linewidth]{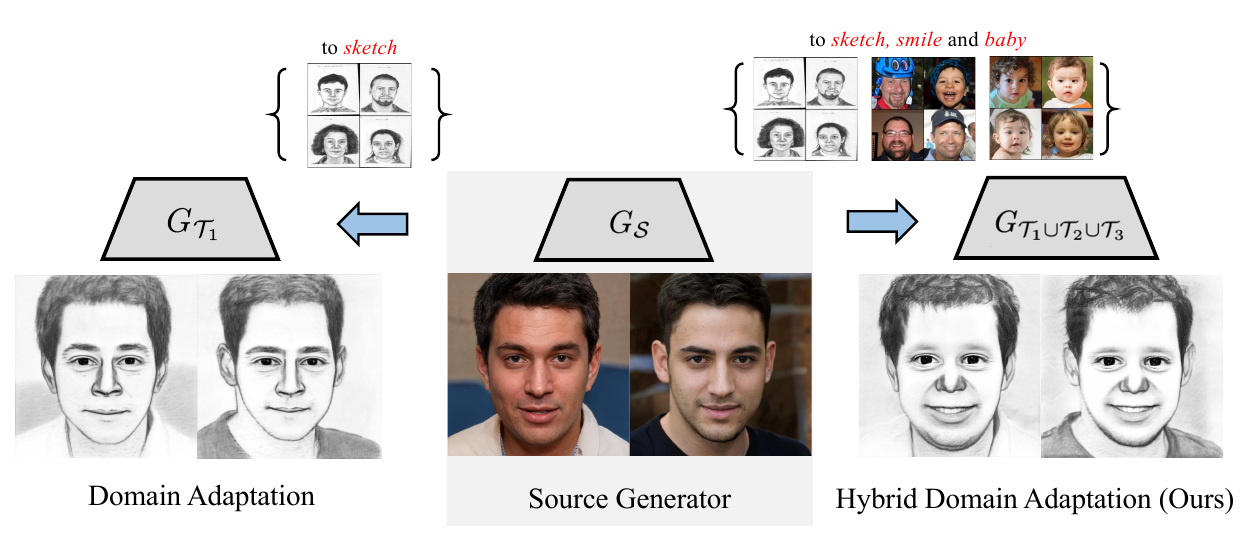}
    \caption{\textbf{(Middle)} Given a source generator ($G_\mathcal{S}$) pre-trained on a large-scale dataset, we propose to adapt it to the new target domain. \textbf{(Left)} \textit{Domain Adaptation} adapts the generator from the source domain (\textit{human}(\textit{sketch}). The adapted model $G_{\mathcal{T}_1}$ captures the target distribution using extremely few-shot references.
   \textbf{(Right)} Our newly introduced \textit{Hybrid Domain Adaptation} aims to adapt to $G_{\mathcal{T}_1\cup \mathcal{T}_2\cup \mathcal{T}_3}$ that generate images with integrated attributes from $\mathcal{T}_1$ (\textit{sketch}), $\mathcal{T}_2$ (\textit{smile}), and $\mathcal{T}_3$ (\textit{baby}), given few-shot references from multiple individual domains.}
    \label{fig:task}
\end{figure}

Few-shot generative \textit{domain adaptation} (DA) aims to adapt a pre-trained image generator~\citep{karras2019style,brock2018large,vahdat2021score,rombach2022high} from the source domain to a new target domain using only few reference images. Existing methods~\citep{mo2020freezeD, li2020fig_EWC, xiao2022few, mondalfew} generally seek to achieve realistic and diverse generation which acquires salient characteristics of the target domain and preserves the variations learned from the source domain. Building upon previous promising progress toward DA, it is straightforward to derive the adapted model given images from \textit{sketch} domain as shown in \cref{fig:task} (Left).
However, a challenge remains when we require to generate images with integrated attributes of \textit{sketch}, \textit{smile}, and \textit{baby}, given real images from individual domains.
Besides, in real-world scenarios, images from the hybrid domain (\eg, a \textit{smiling} \textit{baby} with the style of \textit{sketch}) tend to be more difficult to collect compared with single domain. Under such circumstances, conventional DA becomes less feasible.

Instead of DA, we investigate this novel task -- Few-shot generative \textit{Hybrid Domain Adaptation} (HDA). Given a source generator and few-shot images of several target domains, HDA aims to acquire an adapted image generator that preserves the integrated attributes of all target domains, without overriding the original characteristics. Compared with DA, HDA offers greater flexibility and versatility to adapt generators to more composite and expansive domains. For example, given the few-shot references from the individual \textit{sketch}, \textit{baby} and \textit{smile} domains as shown in \cref{fig:task} (Right), HDA aims to adapt the generator to \textit{sketch-smile-baby} domain. Concurrently, these images retain the primary characteristics of the source domain, which upholds cross-domain consistency. 

Compared with conventional DA, HDA is more challenging in two aspects: (1) Existing DA approaches typically employ the discriminator to discern whether generated images belong to the target domain. However, we only have images sampled from individual domains and lack real images of the hybrid domain, which presents challenges for designing the discriminator-based adaptation framework. (2) Since there are extremely few reference images, the discriminator could easily overfit the easy-to-learn characteristics of certain target domains~\citep{ojha2021fig_cdc}, leading to missing the characteristics from other target domains in the generator.

To solve these issues stemming from the reliance on discriminator, we propose a discriminator-free framework for HDA. Instead of directly distinguishing whether generated images are real or fake, we endeavor to project distinct domains into separate embedding subspaces. Inspired by image classification task~\citep{dosovitskiy2020image, oquab2023dinov2, liu2021swin} that encodes images into high-dimensional embedding space and extract well-separable features, we posit that the pre-trained image encoder intrinsically functions as a hybrid domain classifier. Leveraging this property, we utilize these well-known pre-trained encoders to embed different domains' images into several distinct embedding subspaces.

Employing these well-separable subspaces, we introduce the directional subspace loss comprised of a distance loss and a direction loss to achieve HDA. Specifically, our objective is for generated images to progressively approach all the target subspaces. To this end, we propose the distance loss to minimize the distances from the generated images to all target domains' subspace. Solely employing the distance loss can result in model collapse, since distinct generated images may project onto the same point on the subspace and compromise cross-domain consistency. To preserve more characteristics from the source domain, we further propose the direction loss to guide the adaptation along the perpendicular to subspaces. Additionally, we can easily use our directional subspace loss for single domain adaptation since DA is a special form of HDA.

We validate the effectiveness of our method across multiple target domains.
Compared with other potential approaches for achieving hybrid domain~\citep{wu2023domain, gal2021stylegan}, our method accomplishes the fastest adaptation without training models on
multiple individual domains. For both DA and HDA, our method achieves superior semantic similarity with the target domain, better image fidelity, and more characteristics preservation of the source domain compared with existing approaches. Overall, our contributions are summarized as follows:
\begin{itemize}[itemsep=0pt, parsep=0pt, leftmargin=10pt]
\item We introduce a novel task – Hybrid Domain Adaptation (HDA). Compared with DA, HDA offers greater flexibility and versatility to adapt generators to more composite and expansive domains.
\item We propose a novel discriminator-free framework with directional subspace loss for HDA. Compared with other potential approaches, our methods accomplish fast and versatile adaptation without training models on multiple individual domains.
\item For both DA and HDA, our method surpasses existing methods on the semantic similarity with the target domain, image fidelity, and cross-domain consistency. Qualitative and quantitative results demonstrate the effectiveness of our method.
\end{itemize}

\section{Related Works}
\label{related}
\textbf{Few-shot Generative Domain Adaptation.}
Few-shot generative domain adaptation aims to adapt a pre-trained image generator to a new target domain with a limited number of reference images. Due to the scarcity of training images, existing works~\citep{mo2020freezeD, li2020fig_EWC, ojha2021fig_cdc, zhao2022closer, xiao2022few, mondalfew} often adopt extra regularization terms to avoid overfitting. For example, CDC~\citep{ojha2021fig_cdc} proposes the instance distance consistency loss to preserve the distance between different instances in the source domain.  RSSA~\citep{xiao2022few} proposes a relaxed spatial structural alignment method to preserve the spatial structural information of the source domain.  AdAM~\citep{zhao2022few} proposes Adaptation-Aware kernel Modulation to address general FSIG of different source-target domain proximity. While previous works achieve promising progress toward generative domain adaptation, they rely fundamentally on the discriminator which makes it difficult to handle hybrid domain adaptation.

\textbf{Inference Time Interpolation for Hybrid Domain Adaptation.}
Taking advantage of the disentanglement in the latent space of StyleGAN~\citep{harkonen2020ganspace, shen2021closed, xu2022mind,shen2020interfacegan,wu2020stylespace}, some works~\citep{wu2023domain, nitzan2023domain, gal2021stylegan} are capable of producing images from the hybrid domain through the interpolation technique. For example, DoRM~\citep{wu2023domain} adapts the source generator to each domain individually and then interpolate multiple domain's latent codes at inference time to generate images from the hybrid domain. However, the approach of inference time interpolation needs to train models on multiple individual domains, which necessitates multiple times the model size and training time. In contrast, our method can acquire one model for the hybrid domain in just few minutes.

\textbf{Pre-trained Image Encoder as Implicit Domain Classifier.}
Image encoder has been extensively explored to extract salient and representative features, which typically projects the images into a well-separable embedding space for classification. Recently, beginning with ViT~\citep{dosovitskiy2020image, caron2021emerging, oquab2023dinov2, liu2021swin}, Transformer-based network achieves revolutionary performance on the image classification challenge. To address large variations in the scale of visual entities and the high resolution of pixels in images, Swin~\citep{liu2021swin} proposes a hierarchical Transformer whose representation is computed with shifted windows. On the other hand, self-supervised efforts represented by DINO~\citep{caron2021emerging} and DINOv2~\citep{oquab2023dinov2} have also been devoted to developing effective image encoder. In this paper, we propose to utilize the pre-trained image encoder as an implicit domain classifier to encode the reference images from different domains into the distinct embedding subspace (see \cref{analysis}).

\textbf{Discriminator-Free Domain Adaptation.}
Style-NADA~\citep{gal2021stylegan} proposes text-driven DA and presents a discriminator-free method which utilizes CLIP~\citep{radford2021learning} model to produce the CLIP-space direction. Although the method can also be applied for image-driven domain adaptation, it suffers from the semantic bias stemming from the CLIP encoder and is not guaranteed to converge to the exact realization of the style such as the \textit{sketch} domain. Besides, the directional loss with CLIP cannot handle small attribute edits such as the \textit{sunglasses} domain. Instead, we propose the directional subspace loss with the image encoder pre-trained on large classification dataset, which facilitates the adaptation of small attributes and significantly alleviates the bias. 

\section{Method}
\subsection{Problem Formulation}
Following previous works~\citep{ojha2021fig_cdc, xiao2022few}, we start with a pre-trained StyleGAN2~\citep{karras2019style} generator $G_\mathcal{S}$ that maps from noise $z$ to images in a source domain $\mathcal{S}$, and a set of $N$ target domains $\mathcal{T}_i, i \in \{1, ..., N\}$. 
\textit{Domain adaptation} (DA) finetunes $G_\mathcal{S}$ with few-shot images $\bm{Y}_i$ from each target domain $\mathcal{T}_i$ to yield a generator $G_{\mathcal{T}_i}$. The adapted $G_{\mathcal{T}_i}$ can generate images $\widetilde{\bm{Y}}_i$ similar to domain $\mathcal{T}_i$. Differently, \textit{hybrid domain adaptation} (HDA) aims to acquire a generator $G_\mathcal{T}$ that models a hybrid domain $\mathcal{T} = \cup_{i=1}^{N}\mathcal{T}_i$ and generate images $\widetilde{\bm{Y}}$ with integrated attributes. 
\subsection{Discriminator-free Domain Adaptation}
\label{encoder}
Conventional DA methods commonly incorporate a discriminator to differentiate whether the generated images pertain to the target domain. However, we only have images sampled from individual domains and lack real images of the hybrid domain, which poses challenges in the development of discriminator-based methods. Moreover, owing to the scarcity of reference images~\citep{ojha2021fig_cdc}, the discriminator inevitably exhibits a tendency to overfit to easily learned attributes of certain target domains. Consequently, this leads the generator to ignore other target domains' attributes within the context of HDA.

\begin{figure}
    \centering
\includegraphics[width=0.95\linewidth]{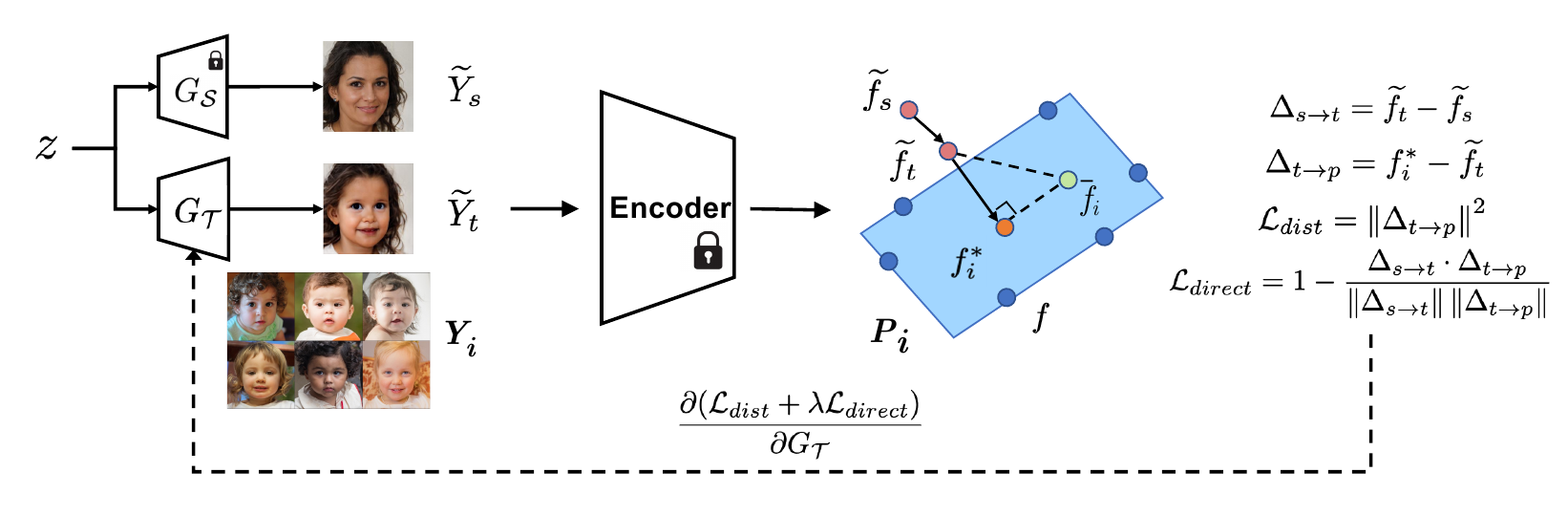}
    \caption{The diagram of our discriminator-free framework with the directional subspace loss $\mathcal{L}_{dist}$ and $\mathcal{L}_{direct}$. $\bm{Y}_i$ is the reference images from the $i$-th domain. $\widetilde{Y}_s$ and $\widetilde{Y}_t$ are images generated by frozen source generator and training target generator. The frozen image encoder extracts features $f$ to constitute the subspace $\bm{P}_i$. After that, we project the target embedding $\widetilde{f}_t$ onto the subspace $\bm{P}_i$ to obtain $f^*_i$. Then we minimize the distance between $\widetilde{f}_t$ and $f^*_i$. Simultaneously, we minimize the angle between $\widetilde{f}_t - \widetilde{f}_s$ and $f^*_i - \widetilde{f}_t$ to guarantee that $\widetilde{f}_t$ moves along the perpendicular from $\widetilde{f}_s$ to $\bm{P}_i$.}
    \label{fig:subspace}
\end{figure}

To address the issues, we propose a discriminator-free framework for HDA. Specifically, we utilize the pre-trained image encoder $E$, such as Swin~\citep{liu2021swin} and Dinov2~\citep{oquab2023dinov2} in image classification task, to encode the few-shot reference images $\bm{Y}_i$ from $\mathcal{T}_i$ into features $\bm{f}_i$:
\begin{equation}
\label{f}
\bm{f}_i = E(\bm{Y}_i), \quad  \overline{f}_i = \frac{1}{|\bm{f}_i|}\sum_{f\in \bm{f}_i} f,
\end{equation}
where $\overline{f}_i$ is the average mean of $\bm{f}_i$ and $|\bm{f}_i|$ is the cardinality of $\bm{f}_i$ that equals to the number of images in $\mathcal{T}_i$. As shown in \cref{fig:subspace}, the features $f$ in $\bm{f}_i$ constitute a separable embedding subspace $\bm{P}_i$ (see \cref{analysis}).

\subsection{Directional Subspace Loss}
\label{sec:subspace}
Then we introduce a directional subspace loss for the adaptation. Technically, we encode the images generated by the frozen source generator $G_\mathcal{S}$ and the adapted generator $G_\mathcal{T}$ with the same noise $z$. 
\begin{equation}
\begin{gathered}
\widetilde{Y}_s = G_\mathcal{S}(z), \quad \widetilde{Y}_t = G_\mathcal{T}(z)
\\
\widetilde{f}_s = E(\widetilde{Y}_s), \quad \widetilde{f}_t = E(\widetilde{Y}_t).
\end{gathered}
\end{equation}
Our objective is for generated images to progressively approach the target subspace. Inspired by \citep{simon2020adaptive}, we propose the distance loss from the generated images to the target subspace. Concretely, we first project the embedding $\widetilde{f}_t$ onto the subspace $\bm{P}_i$ to get the projected point $f^*_i$:
\begin{equation}
f^*_i = \bm{M}_i\bm{M}_i^T(\widetilde{f}_t - \overline{f}_i) + \overline{f}_i,
\end{equation}
where $\bm{M}_i$ is the projection matrix computed from SVD factorization of all elements in $\bm{f}_i$.
We then minimize the distance between $\widetilde{f}_t$ and subspace $\bm{P}_i$ spanned by $\bm{f}_i$:
\begin{equation}
\label{s}
\begin{aligned}
\mathcal{L}_{dist} = dist(\widetilde{f}_t, \bm{P}_i) = \left\|f^*_i - \widetilde{f}_t \right\|^2.
\end{aligned}
\end{equation}
However, solely employing $\mathcal{L}_{dist}$ can still result in model collapse, \ie, distinct generated images may project onto the same point on $\bm{P}_i$ which compromises cross-domain consistency. To preserve more characteristics from $\widetilde{Y}_s$, we propose to enforce the collinearity between $\widetilde{f}_s$, $\widetilde{f}_t$, and $f^*_i$ such that $f^*_i$ becomes the closest point to $\widetilde{f}_s$ on the subspace $\bm{P}_i$. To this end, we minimize the angle between $\widetilde{f}_t - \widetilde{f}_s$ and $f^*_i - \widetilde{f}_t$ to guarantee that $\widetilde{f}_t$ moves along the perpendicular from $\widetilde{f}_s$ to the subspace $\bm{P}_i$:
\begin{equation}
\begin{gathered}
\Delta_{s\rightarrow t} = \widetilde{f}_t - \widetilde{f}_s, \quad \Delta_{t\rightarrow p} = f^*_i - \widetilde{f}_t~, \\ 
\mathcal{L}_{direct} = 1 - \frac{\Delta_{s\rightarrow t} \cdot \Delta_{t\rightarrow p}}{\left\|\Delta_{s\rightarrow t}\right\|\left\|\Delta_{t\rightarrow p}\right\|}~,
\end{gathered}
\end{equation}
Besides, to alleviate the biases introduced by individual image encoder, we employ the ensemble method that exploits several image encoders during the adaptation:
\begin{equation}
\begin{aligned}
\mathcal{L}_{\mathrm{DA}} = \mathbb{E}_{z\sim p(z)}\sum_{E \in \bm{E}}(\mathcal{L}_{dist} + \lambda\mathcal{L}_{direct}),
\end{aligned}
\label{eq:da}
\end{equation}
 where $\bm{E}$ is the set of pre-trained image encoders and $\lambda$ is the balancing factor.
\subsection{Hybrid Domain Adaptation}
Naturally, we can extend the directional subspace loss for HDA, since DA is a special form of HDA. For the distance loss, we reduce the distances between $\widetilde{f}_t$ and all target subspaces to ensure that the adapted generator owns their attributes:
\begin{equation}
\mathcal{L}_{dist}' = \sum_{i=1}^N\alpha_i\left\|f^*_i - \widetilde{f}_t \right\|^2,
\label{hda-dist}
\end{equation}
where $\alpha_i$ is pre-defined domain coefficient for the $i$-th domain.

On the other hand, we also extend the direction loss for HDA to preserve the characteristics of the source domain $\widetilde{f}_s$. Specifically, we minimize the angle between $\widetilde{f}_t - \widetilde{f}_s$ and the weighted sum of $f^*_i - \widetilde{f}_t$ (the perpendicular  from $\widetilde{f}_t$ to the $i$-th subspace): 
\begin{equation}
\begin{gathered}
\Delta_{t\rightarrow p}' = \sum_{i=1}^N\alpha_i(f^*_i - \widetilde{f}_t)~, \\
\mathcal{L}_{direct}' = 1 - \frac{\Delta_{s\rightarrow t} \cdot \Delta_{{t\rightarrow p}}'}{\|\Delta_{s\rightarrow t}\|\|\Delta_{{t\rightarrow p}}'\|}.~
\end{gathered}
\label{hda-direct}
\end{equation}

Overall, our final objective for HDA is 
\begin{equation}
\begin{gathered}
\mathcal{L}_\mathrm{HDA} = \mathbb{E}_{z\sim p(z)}\sum_{E \in \bm{E}}(\mathcal{L}_{dist}' + \lambda\mathcal{L}_{direct}'),
\end{gathered}
\label{eq:hda}
\end{equation}

\begin{figure}[t]
    \centering
\subfigure{
    \begin{minipage}[t]{1\textwidth}
    \centering
    \includegraphics[width=0.9\textwidth]{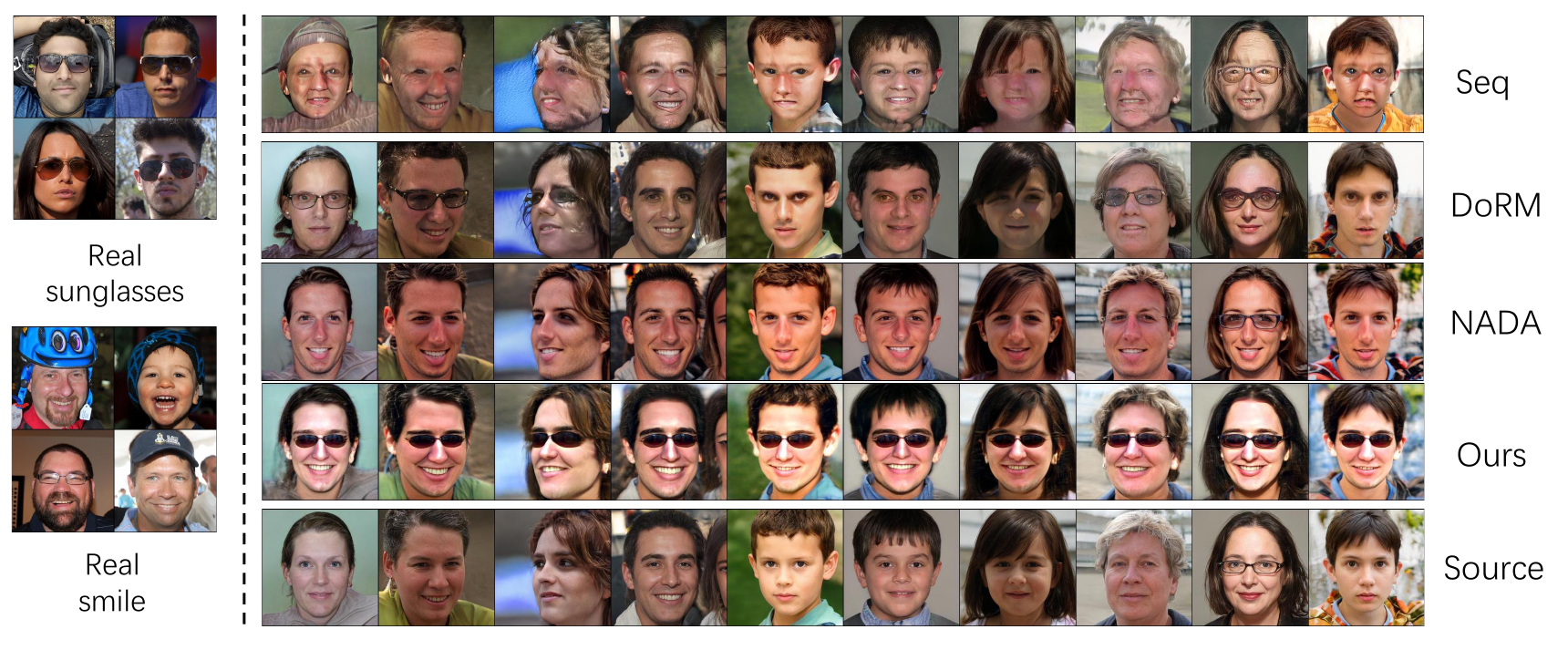}
    \end{minipage}
\label{fig:smile_sunglass}
}
 \\
 \vspace{-0.3cm}
\subfigure{
    \begin{minipage}[t]{1\textwidth}
    \centering
        \includegraphics[width=0.9\textwidth]{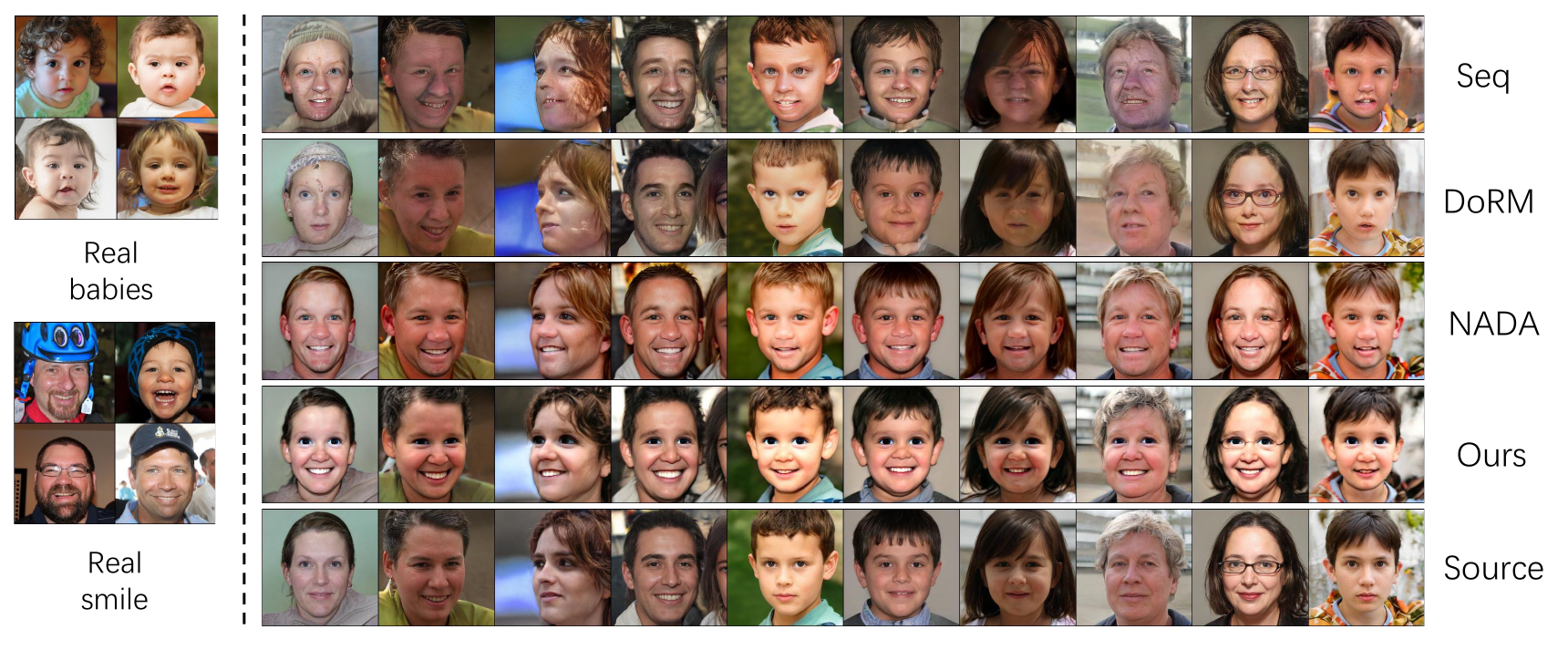} 
    \end{minipage}
    \label{fig:baby_smile}
}
\vspace{-0.5cm}
\caption{Qualitative results on 10-shot HDA of \textit{smile-sunglasses} and \textit{baby-smile}.}
\label{fig:HDA}
\vspace{-0.3cm}
\end{figure}

\begin{figure}
    \centering
    \includegraphics[width=0.80\linewidth]{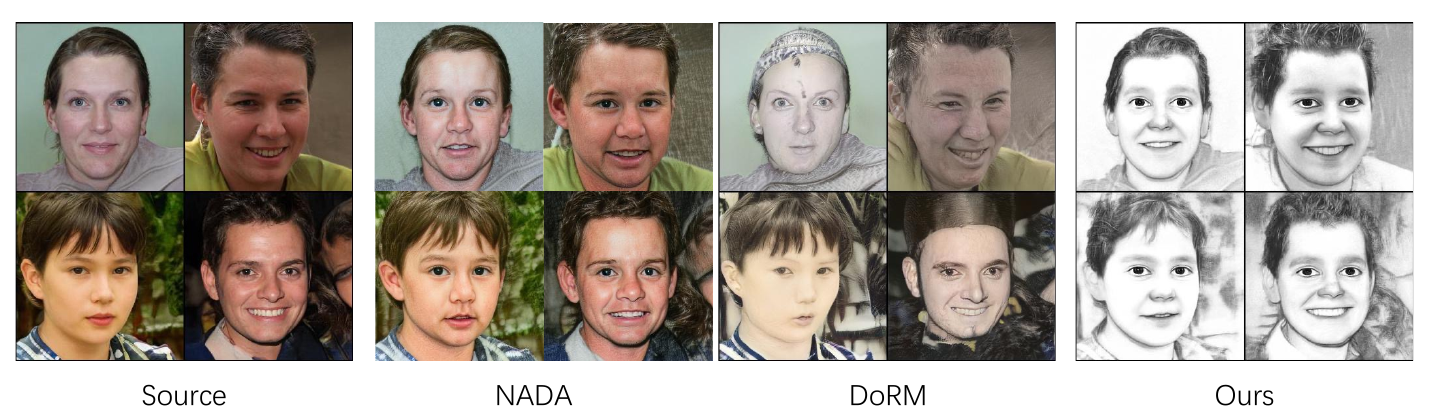}
    \vspace{-0.3cm}
    \caption{Qualitative results on 3 domain 10-shot HDA of \textit{sketch-smile-baby}.}
    \label{fig:3domain}
    \vspace{-0.5cm}
\end{figure}
\vspace{-0.2cm}
\section{Experiments}
\subsection{Datasets and Metrics}
\textbf{Datasets:}
\label{dataset}
Following previous literature, we consider Flickr-Faces-HQ (FFHQ)~\citep{karras2019style} as one of the source domains and adapt to the combination of the following individual target domains (a) FFHQ-\textit{baby}, (b) FFHQ-\textit{sunglasses}, (c) \textit{sketch}, and (d) FFHQ-\textit{smile}. As in previous works, all our experiments use 10 randomly sampled targets for each domain. Unless stated otherwise, we operate on $256 \times 256$ resolution images for both the source and target domains.

\textbf{Metrics:}
\label{metric}
A key difference between the proposed HDA task against previous DA~\citep{ojha2021fig_cdc, mondalfew, nitzan2023domain} is that there are no real images in the hybrid target domains. In terms of the quantitative evaluation, we focus on these evaluation metrics in our experiments: (1) CLIP-Score~\citep{nitzan2023domain, radford2021learning} measures the compatibility of image-text pairs, which can be thought of as the semantic similarity with the target domain. (2) The Inception Score (IS) is used to assess the fidelity of images, which is practical in the few shot setting~\citep{xiao2022few}. (3) Identify similarity (ID) measures cross-domain consistency where the adapted images retain consistency with their corresponding source images in domain-invariant aspects like pose and identity, following~\citep{zhang2022generalized, wu2023domain}.

\subsection{Methodology and Baselines}
\textbf{Methodology:} Following previous domain adaptation (DA) works~\citep{ojha2021fig_cdc, xiao2022few, mondalfew, zhao2022closer}, we apply our method on the StyleGAN2~\citep{karras2020analyzing} architecture. To show the flexibility of our method, we conduct the experiments on both DA and HDA. As described in \cref{encoder}, we utilize Swin~\citep{liu2021swin} and Dinov2~\citep{oquab2023dinov2} as the image encoder, which are pre-trained on ImageNet 22k dataset~\citep{deng2009imagenet}.

\textbf{Baselines:}
\label{baseline} 
To demonstrate the effectiveness of our method, we compare our method against the following baselines that could potentially achieve hybrid domain:
(1) Sequential: sequentially adapted to individual target domains to achieve hybrid domain; (2) DoRM: Domain Remodulation~\citep{wu2023domain} for hybrid domain as discussed in \cref{related}; (3) NADA: Following NADA~\citep{gal2021stylegan}, we interpolate the model parameters in different domains to achieve hybrid domain. Additionally for DA, we compare ours with CDC~\citep{ojha2021fig_cdc}, DoRM~\citep{wu2023domain}, RSSA~\citep{xiao2022few}, AdAM~\citep{zhao2022few}, and NADA~\citep{gal2021stylegan}.  

\begin{table*}[t]
\small
  \centering
  \setlength{\tabcolsep}{4pt}
  \renewcommand\arraystretch{1.3}
\begin{tabular}{cccccccccc}
\toprule
\multicolumn{1}{c}{\multirow{2}{*}{Method}}& \multicolumn{3}{c}{\textit{smile-sunglass}} & \multicolumn{3}{c}{\textit{baby-smile}} &
\multicolumn{3}{c}{\textit{baby-sketch}}  
\\
\cmidrule(lr){2-4} \cmidrule(lr){5-7} \cmidrule(lr){8-10}
    \multicolumn{1}{c}{}
    & {\scriptsize{CS} ($\uparrow$)}
    & {\scriptsize{IS} ($\uparrow$)} 
    & {\scriptsize{ID} ($\uparrow$)}   
    & {\scriptsize{CS} ($\uparrow$)}
    & {\scriptsize{IS} ($\uparrow$)} 
    & {\scriptsize{ID} ($\uparrow$)}
    & {\scriptsize{CS} ($\uparrow$)}
    & {\scriptsize{IS} ($\uparrow$)} 
    & {\scriptsize{ID} ($\uparrow$)}
    \\ \midrule
  Seq    
 &17.92	&1.82	&0.261
&17.36	&1.7	&0.33 
 &22.69	&2.28	&0.198  
 \\ 
  DoRM  
 &18.99	&1.94	&0.312  
&16.38	&2.12	&0.28
 &17.46	&2.69	&0.21
 \\ 
 NADA 
  &20.23	&1.34	&0.27
 &20.45	&1.68	&\textbf{0.34} 
  &22.06	&2.61	&0.18  
 \\ 
 Ours   
&\textbf{25.65}	&\textbf{2.4}	&\textbf{0.373}
&\textbf{23.08}	&\textbf{2.94}	&\textbf{0.34}
&\textbf{24.91}	&\textbf{2.97}	&\textbf{0.223}
 \\ 
 \bottomrule

\end{tabular}
\centering
\caption{Quantitative evaluations on 10-shot HDA including CLIP-Score (CS), IS, and ID discussed in \cref{metric}. Note that $\uparrow$ indicates higher is better.}
\label{tab:hda}
\end{table*}

\begin{table*}[t]
\small
  \centering
  \setlength{\tabcolsep}{4pt}
  \renewcommand\arraystretch{1.3}
\begin{tabular}{ccccccccc}
\toprule
\multicolumn{1}{c}{\multirow{2}{*}{Method}}&
\multicolumn{1}{c}{\multirow{2}{*}{\textit{Inter.}}}&
\multicolumn{1}{c}{\multirow{2}{*}{\textit{Dis. free}}}&
\multicolumn{2}{c}{2-domain}& 
\multicolumn{2}{c}{5-domain} &
\multicolumn{2}{c}{10-domain}  
\\
\cmidrule(lr){4-5} \cmidrule(lr){6-7} \cmidrule(lr){8-9} 
    \multicolumn{1}{c}{}
    &\multicolumn{1}{c}{}
    &\multicolumn{1}{c}{}
    & {\scriptsize{size} ($\downarrow$)}
    & {\scriptsize{time} ($\downarrow$)} 
    & {\scriptsize{size} ($\downarrow$)} 
    & {\scriptsize{time} ($\downarrow$)}    
    & {\scriptsize{size} ($\downarrow$)} 
    & {\scriptsize{time} ($\downarrow$)}\\
    \midrule 
 Seq & &
&24M &240min &24M & 600min &24M &1200min
\\
 DoRM &\checkmark & 
 &30M &240min
 &54M &600min
 &84M &1200min  
 \\ 
NADA &\checkmark &\checkmark 
&48M &6min &120M & 15min &240M &30min
\\
 \midrule
 Ours  & &\checkmark
&\textbf{24M} &\textbf{3min}
&\textbf{24M} &\textbf{3min}
&\textbf{24M} &\textbf{3min}

 \\ 
 \bottomrule

\end{tabular}
\centering
\caption{Comparison of model size and training time. Note that \textit{Inter.} and \textit{Dis. free} indicate the inference time interpolation and discriminator-free methods.}
\label{tab:time}%
\vspace{-0.2cm}
\end{table*}

\begin{figure}[t]
    \centering
\subfigure{
    \begin{minipage}[t]{1\textwidth}
    \centering
    \includegraphics[width=0.9\textwidth]{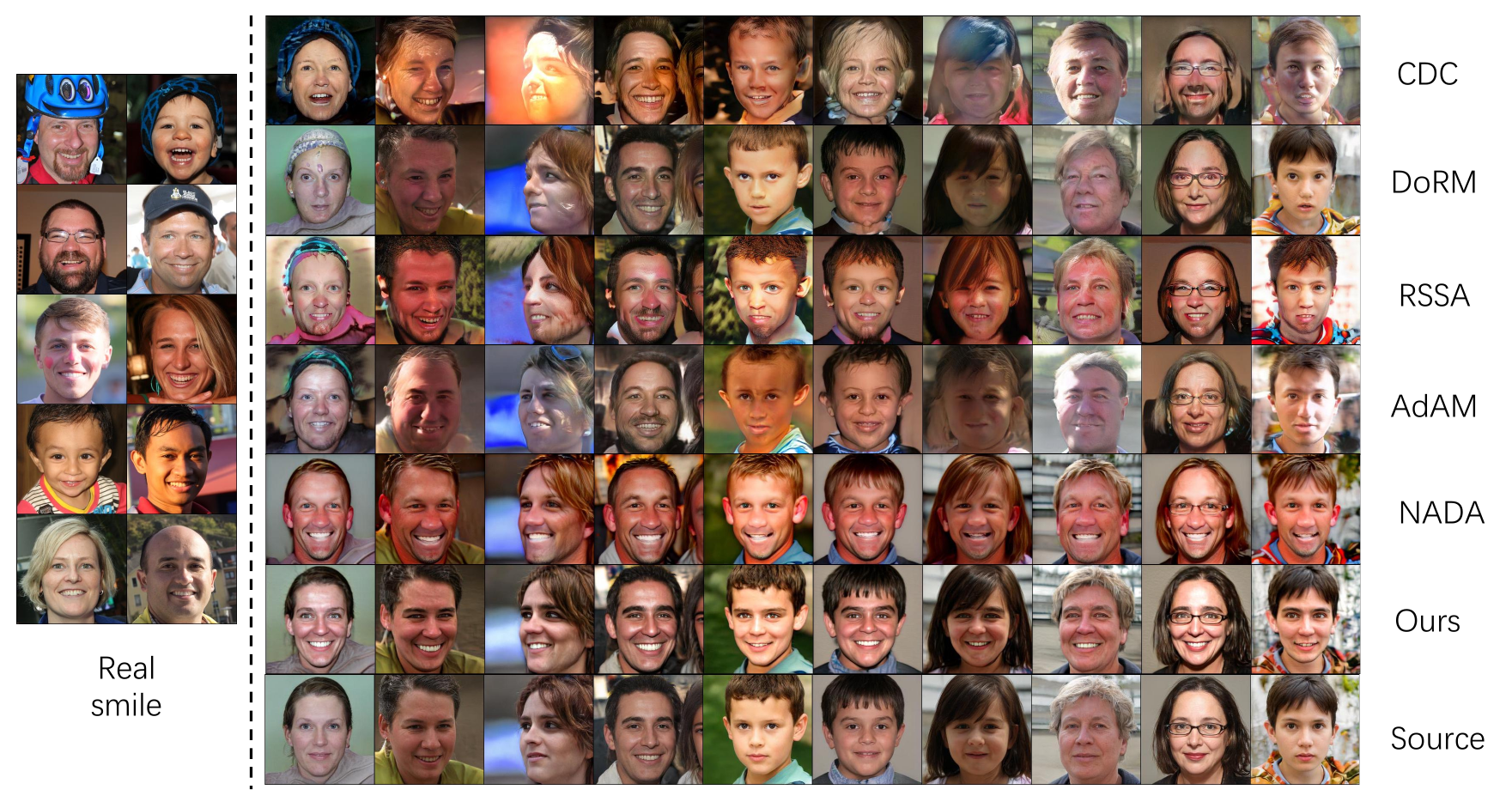}
    \end{minipage}
\label{fig:smile}
}
\\
 \vspace{-0.3cm}
\subfigure{
    \begin{minipage}[t]{1\textwidth}
    \centering
        \includegraphics[width=0.9\textwidth]{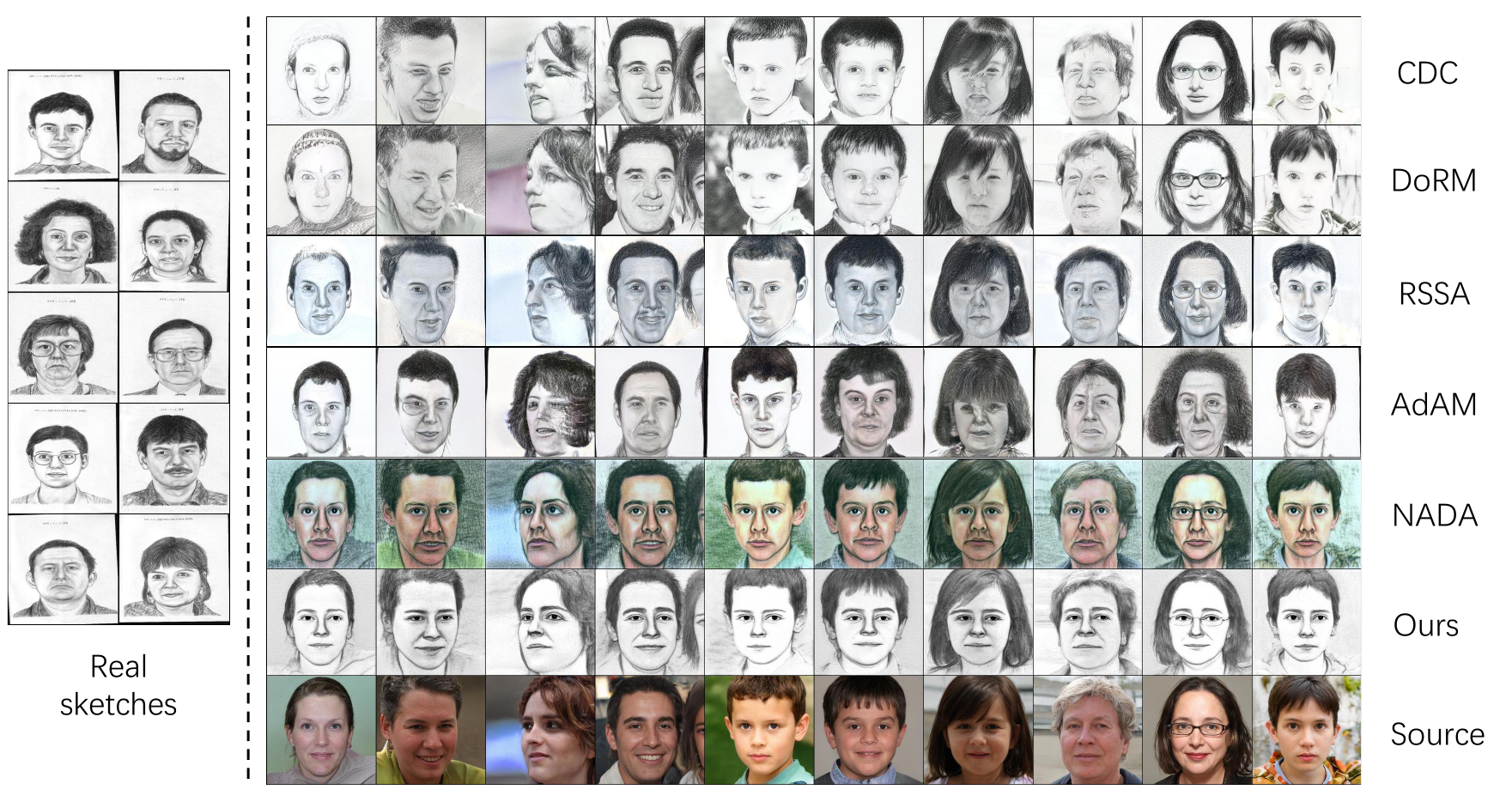} 
    \end{minipage}
    \label{fig:sketch}
}
\vspace{-0.5cm}
    \caption{Qualitative results on 10-shot DA of \textit{smile} and \textit{sketch}.}
    \label{fig:DA}
    \vspace{-0.2cm}
\end{figure}

\begin{table*}[t]
\small
  \centering
  \setlength{\tabcolsep}{4pt}
  \renewcommand\arraystretch{1.3}
\begin{tabular}{ccccccccccccc}
\toprule
\multicolumn{1}{c}{\multirow{2}{*}{Method}}& \multicolumn{3}{c}{\textit{baby}} & \multicolumn{3}{c}{\textit{smile}} &
\multicolumn{3}{c}{\textit{sunglasses}}&
\multicolumn{3}{c}{\textit{sketch}}  
\\
\cmidrule(lr){2-4} \cmidrule(lr){5-7} \cmidrule(lr){8-10} \cmidrule(lr){11-13}
    \multicolumn{1}{c}{}
    & {\scriptsize{CS} ($\uparrow$)}
    & {\scriptsize{IS} ($\uparrow$)} 
    & {\scriptsize{ID} ($\uparrow$)}   
    & {\scriptsize{CS} ($\uparrow$)}
    & {\scriptsize{IS} ($\uparrow$)} 
    & {\scriptsize{ID} ($\uparrow$)}
    & {\scriptsize{CS} ($\uparrow$)}
    & {\scriptsize{IS} ($\uparrow$)} 
    & {\scriptsize{ID} ($\uparrow$)}
    & {\scriptsize{CS} ($\uparrow$)}
    & {\scriptsize{IS} ($\uparrow$)} 
    & {\scriptsize{ID} ($\uparrow$)}\\ \midrule
 CDC  
 &17.84 &2.22 &0.326  
 &18.93 &1.22 &0.479
  &22.19 &2.33 &0.318
  &20.83 &2.12 &0.214
 \\ 
 DoRM
 &17.10 &2.30 &0.335
 &17.10 &2.18 &0.490 
 &22.26 &2.59 &0.389 
 &19.42 &2.06 &0.365
  \\
  RSSA 
&18.42 &2.57 &0.238 
&18.18 &2.20 &0.410
 &21.63 &2.52 &0.256 
 &21.40 &1.66 &0.231
 \\ 
  AdAM
   &21.18 &2.12 &0.210
 &19.79 &1.61 &0.280 
  &22.50 &2.50 &0.197 
  &21.18 &2.22 &0.109
  \\ 
  NADA
   &21.52	&3.36	&0.182
 &19.95	&1.18	&0.305 
  &20.95	&1.51	&0.233 
  &20.9	&1.86	&0.335
 \\ 
 Ours   
&\textbf{21.92} &\textbf{3.71} &\textbf{0.353}
&\textbf{20.38} &\textbf{2.23} &\textbf{0.587}
&\textbf{24.13} &\textbf{2.67} & \textbf{0.414}
&\textbf{22.83} &\textbf{2.39} &\textbf{0.417}
 \\ 
 \bottomrule

\end{tabular}
\centering
\caption{Quantitative evaluations on 10-shot DA.}
\label{tab:da}%
\end{table*}

\subsection{Results on Hybrid Domain Adaptation}
\textbf{Qualitative Results:}
\cref{fig:HDA} shows the qualitative result of the baseline and our method on HDA, all of which start from the same source domain FFHQ~\citep{karras2019style} to the combinations of individual \textit{baby}, \textit{sketch}, \textit{smile} and \textit{sunglasses} domains, respectively. For naive Sequential learning, we find that it suffers from catastrophic forgetting~\citep{mccloskey1989catastrophic} and tends to generate images of the last adapted domain, which fails to generate desired images (\eg, \textit{baby with the style of sketch}). 
DoRM~\citep{wu2023domain} are unstable to generate images of the hybrid domain (\eg, the images of \textit{smile-sunglasses} lack the attribute of \textit{smile} and lose the characteristics of the source domain), since the simple interpolation of latent codes at inference time is not enough to semantically align multiple domains. NADA could somehow generate images with integrated attributes, but it exhibits significant bias stemming from the CLIP model. For instance, the sketch domain deviates stylistically from the reference images, and it fails to learn the sunglasses domain.
In contrast, our method exhibits desirable performance and shows its stable capability to generate hybrid target images and preserve the characteristics of the source domain. 
Furthermore, we also conduct experiments on the hybrid of three target domains \textit{sketch-smile-baby} in \cref{fig:3domain} where our generated images blend the domain-specific attributes and preserve the primary characteristics of the source domain, which surpass all the baseline methods. 
More results for HDA are included in Appendix.

\begin{figure}
\begin{minipage}[t]{.66\textwidth}
    
    \flushright
\subfigure{

    \begin{minipage}[b]{0.9\textwidth}
        \includegraphics[width=1\textwidth]{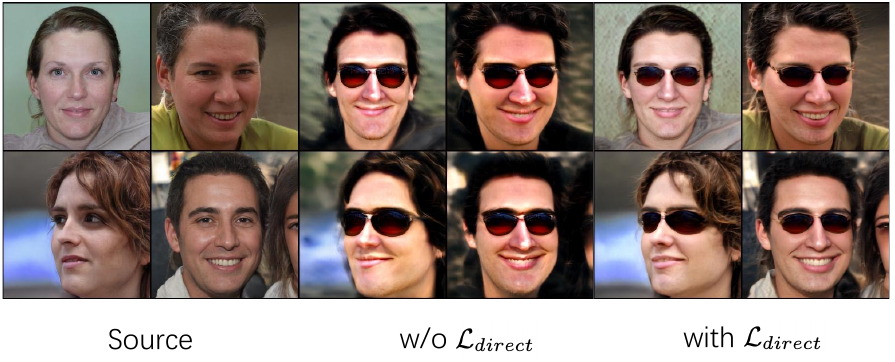}
    \end{minipage}
}
 \end{minipage}%
\begin{minipage}[t]{.33\textwidth}     
\subfigure{
    \begin{minipage}[b]{0.9\textwidth}
        \includegraphics[width=1\textwidth]{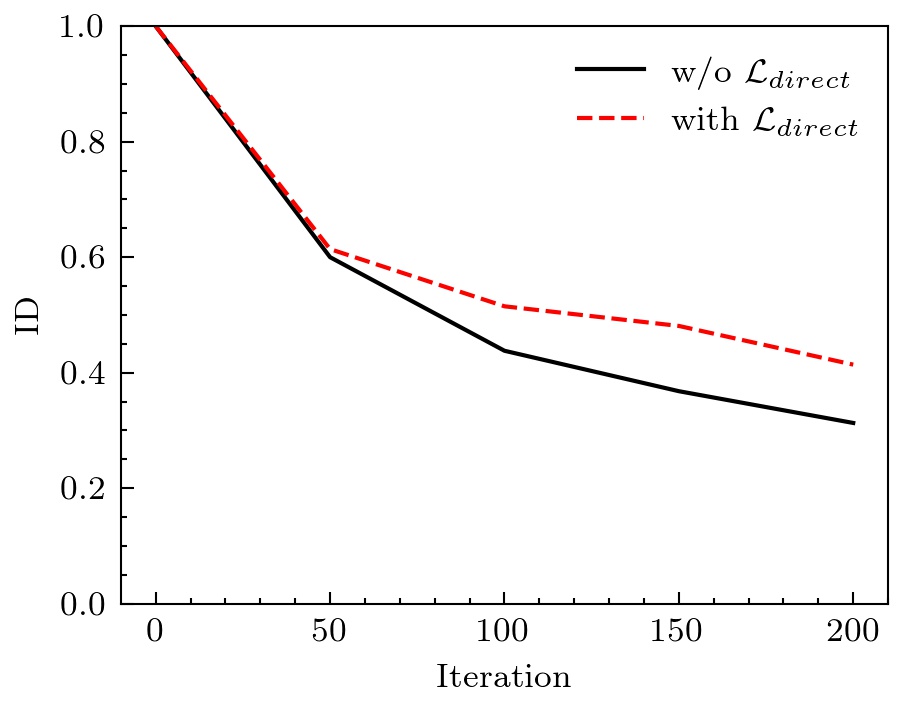} 
    \end{minipage}
}
 \end{minipage}
 \vspace{-0.2cm}
 	\caption{Qualitative ablation for $\mathcal{L}_{direct}$ on \textit{sunglasses}.
  }
  \vspace{-0.2cm}
	\label{fig:ablation}
\end{figure}

\begin{figure}[t]
    \centering
    \includegraphics[width=0.9\linewidth]{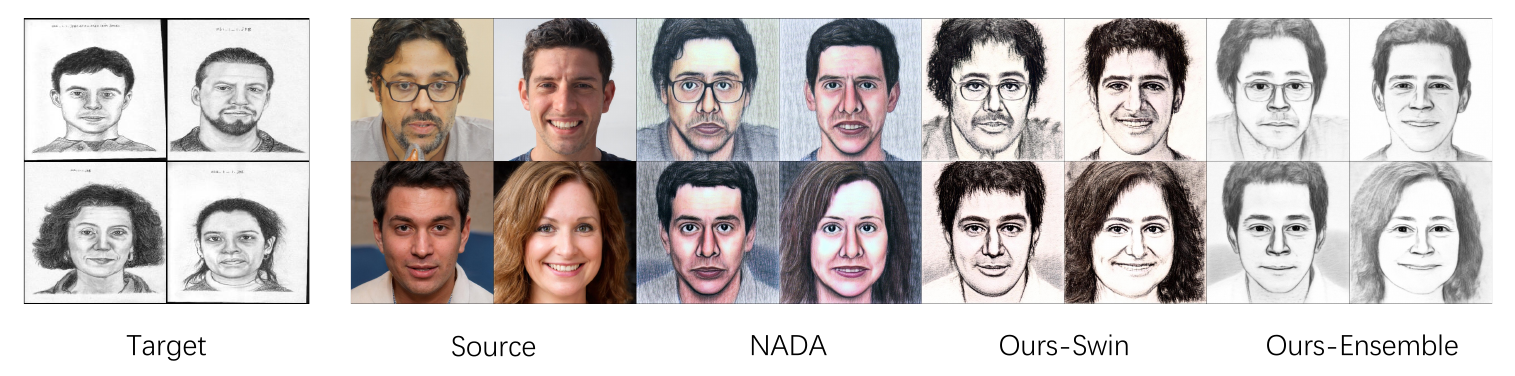}
    \vspace{-0.2cm}
    \caption{Qualitative ablation for the ensemble on \textit{sketch}.}
    \label{fig:ensemble}
    \vspace{-0.2cm}
\end{figure}

\textbf{Quantitative Results:}
To quantify the quality and diversity of the generated images, we evaluate all methods with CLIP-Score, IS, and ID respectively. As shown in \cref{tab:hda}, our method achieves best scores for all metrics. Especially for CLIP-Score, our method significantly outperforms other methods, indicating that generated images with our method effectively integrate the multiple attributes from distinct domains. Furthermore, our method achieves better IS and ID, indicating that generated images has higher fidelity and preserve more characteristics of the source domain.

Additionally, we calculate the model size and training time to verify the efficiency of our method as shown in \cref{tab:time}. Compared with the discriminator-based baselines, our discriminator-free method obviates adversarial training, thereby saving substantial training time. Furthermore, compared with interpolation-based approaches at inference time, our method saves multiple times the model size and training time, completing the adaptation in just three minutes.

\subsection{Results on Single Domain Adaptation}
\textbf{Qualitative Results:}
To show the flexibility of our proposed method, we also conduct the experiments on DA, as shown in \cref{fig:DA}. Similar to the performance on HDA, our method surpasses all baseline approaches. Specifically, previous methods exhibit severe overfitting to the few-shot reference images, inadequately preserving source domain attributes, while also demonstrating relatively inferior image fidelity. While NADA demonstrates cross-domain consistency to some extent, it exhibits significant bias stemming from the CLIP encoder. In contrast, our method is capable of accurately acquiring the visual attributes of target domain while retaining strong fidelity and cross-domain consistency.

\textbf{Quantitative Results:}
To demonstrate the effectiveness of our method on DA, we evaluate all the methods with CLIP-Score, IS, and ID respectively. As reported in \cref{tab:da}, our method surpasses all the baselines. Concretely, our method achieves higher semantic similarity with the target domain in CLIP-Score. Furthermore, it outperforms priors on IS and ID, indicating that our generated images have better fidelity and cross-domain consistency.

\begin{wraptable}{r}{0.55\textwidth}
\small
  \centering
  \setlength{\tabcolsep}{4pt}
  \renewcommand\arraystretch{1.0}
\begin{tabular}{cccccc}
\toprule
\multicolumn{1}{c}{\multirow{2}{*}{$\mathcal{L}_{dist}$ }}& \multicolumn{1}{c}{\multirow{2}{*}{$\mathcal{L}_{direct}$ }}& \multicolumn{2}{c}{\textit{sunglass}} & \multicolumn{2}{c}{\textit{baby}} 
\\
\cmidrule(lr){3-4} \cmidrule(lr){5-6} 
    \multicolumn{1}{c}{}
    & \multicolumn{1}{c}{}
    & {\scriptsize{IS} ($\uparrow$)}
    & {\scriptsize{ID} ($\uparrow$)}
    & {\scriptsize{IS} ($\uparrow$)}
    & {\scriptsize{ID} ($\uparrow$)}    
    \\ \midrule  
 \checkmark&
 &1.91	&0.249
 &2.56	&0.178 
 \\ 
   &\checkmark
 &2.61	& 0.372
 &3.5 &0.304
  \\ 
 \checkmark  & \checkmark
 &\textbf{2.67}	&\textbf{0.414} 
 &\textbf{3.71}	&\textbf{0.353} 
 \\ \bottomrule
\end{tabular}
\centering
\caption{Ablation for $\mathcal{L}_{dist}$ and $\mathcal{L}_{direct}$ on 10-shot DA.}
\label{tab:sub}
\end{wraptable}
\subsection{Ablation Study}
First, we conduct the ablation study to verify the effectiveness of our proposed directional subspace loss. As shown in \cref{fig:ablation}, $\mathcal{L}_{direct}$ greatly improves the cross-domain consistency (\eg, clothes and hair). Consistently, we observe that the adaptation with our directional loss maintains higher ID scores throughout training. Quantitatively, we report the ablation for $\mathcal{L}_{dist}$ and $\mathcal{L}_{direct}$ in \cref{tab:sub}. It can be clearly observed that the $\mathcal{L}_{dist}$ and $\mathcal{L}_{direct}$ significantly improve image fidelity and cross-domain consistency. Besides, we report the qualitative ablation for the encoders ensemble in \cref{fig:ensemble}. Our method with Swin~\citep{liu2021swin} significantly alleviates the bias compared with NADA~\citep{gal2021stylegan} and the ensemble technique further help to converge to the exact realization of the style of the \textit{sketch} domain.

\section{Conclusion \& Limitation}
\textbf{Conclusion.}
In this paper, we propose a new task -- Hybrid Domain Adaptation (HDA), which aims to adapt a pre-trained generator to the hybrid target domain to generate images with integrated attributes. To achieve this, we introduce a novel discriminator-free framework, which utilizes the pre-trained image encoder to encode the reference images from different domains into separate embedding subspaces. Furthermore, we introduce a directional subspace loss to guide the adaptation and retain cross-domain consistency. We believe that our work is an important step towards few-shot HDA since we have demonstrated that the source generator can be effectively adapted to a hybrid domain without accessing the real images from the hybrid domain.

\textbf{Limitation.}
While our method realizes HDA and achieves better superior metrics compared with state-of-the-art DA methods, it also has the limitation. We observe that for hybrid domain adaptation, the weights of different domain in our loss are not entirely equivalent to the proportionality of visual characteristics, which needs empirical adjustment. Nevertheless, we believe that some solutions could be integrated into our HDA pipeline to facilitate the alignment in the future.

\bibliography{main}
\bibliographystyle{main}

\appendix
\section{Appendix}
\subsection{Embedding space analysis}
\label{analysis}
\begin{figure}[t]
\begin{minipage}[t]{.5\textwidth}  
\centering
\subfigure[Swin]{
    \begin{minipage}[b]{1\textwidth}
        \includegraphics[width=1\textwidth]{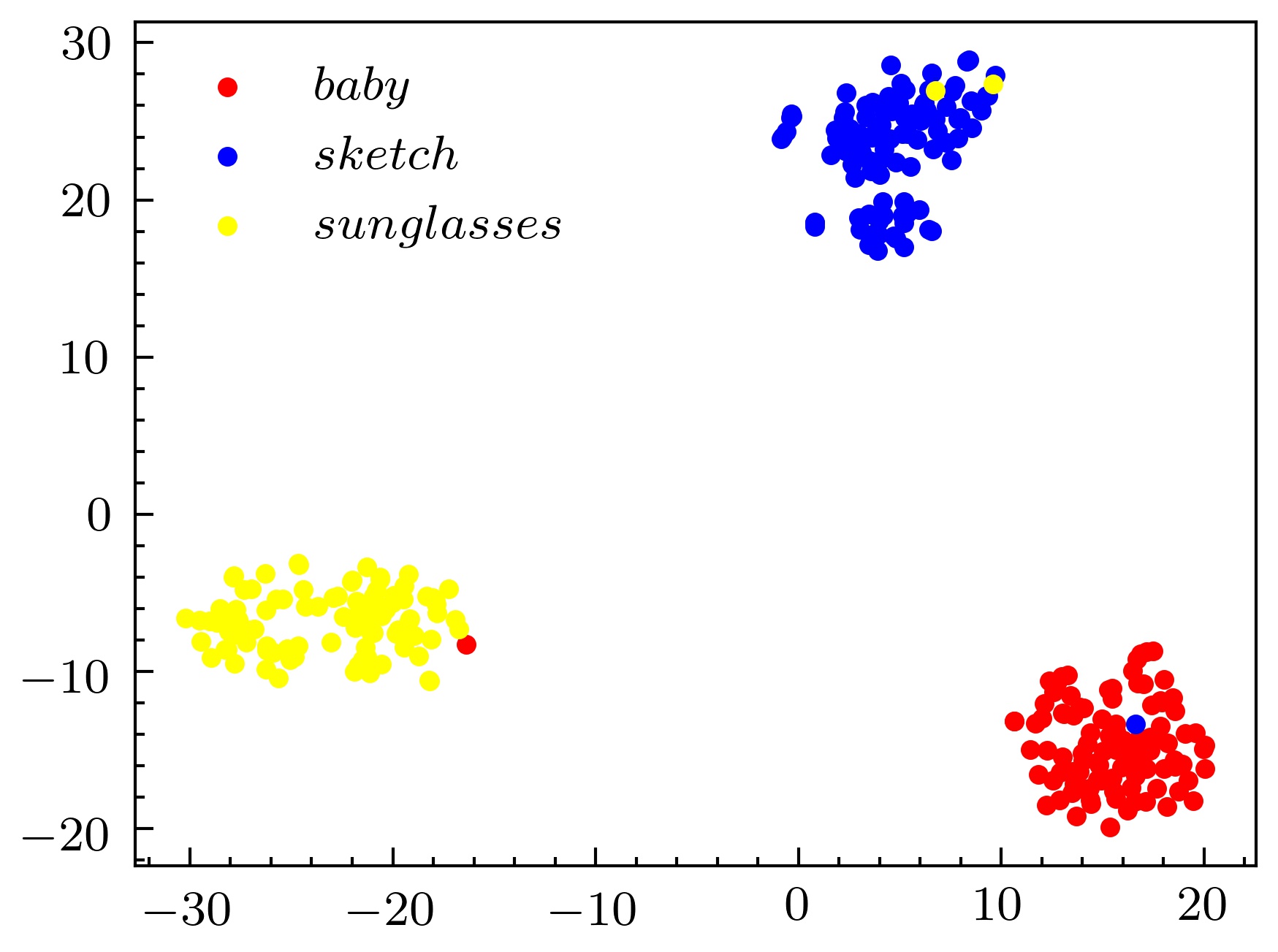} 
    \end{minipage}
    \label{fig:swin}
}
 \end{minipage}%
\begin{minipage}[t]{.5\textwidth}     
\subfigure[Dinov2]{
    \begin{minipage}[b]{1\textwidth}
        \includegraphics[width=1\textwidth]{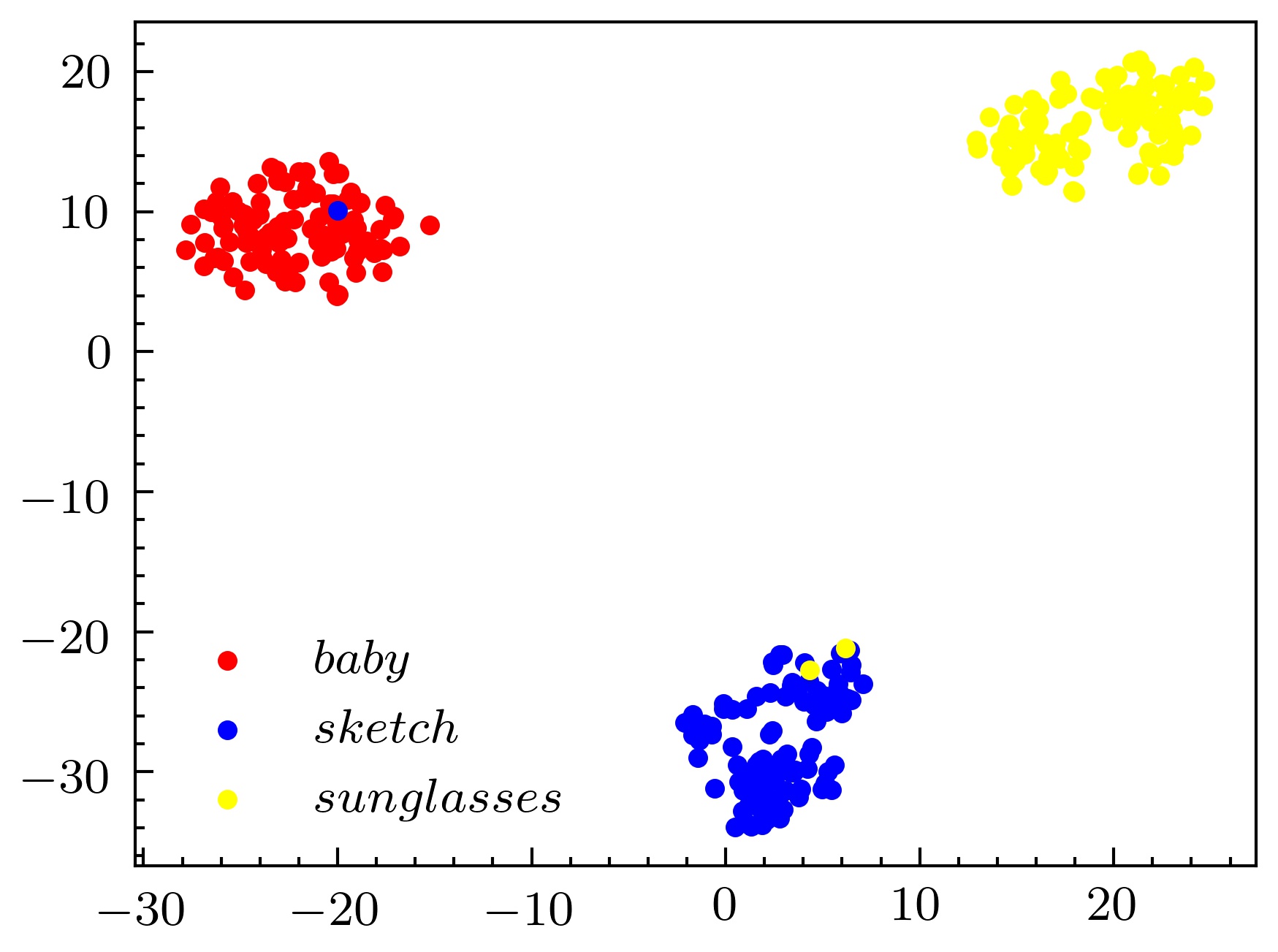} 
    \end{minipage}
    \label{fig:dino}
}
 \end{minipage}
 	\caption{T-SNE visualization of the embedding space encoded by Swin~\citep{liu2021swin} and Dinov2~\citep{oquab2023dinov2} for images from \textit{baby}, \textit{sketch}, and \textit{sunglasses} domain.
  }
	\label{fig:tsne}
\end{figure}

Since the image encoder in image classification typically projects the images into a well-separable embedding space, such as Swin~\citep{liu2021swin} and Dinov2~\citep{oquab2023dinov2}, we can employ the separable subspaces to achieve HDA. To validate that, we visualize the embeddings of real images from multiple domains by t-SNE~\citep{van2008visualizing}. As shown in \cref{fig:tsne}, distinct subspaces from different domains manifest separation, which facilitates us to adapt the generator to the target subspace corresponding to the target domain.

\begin{figure}[t]
    \centering
\subfigure{
    \begin{minipage}[t]{1\textwidth}
    \centering
    \includegraphics[width=0.9\textwidth]{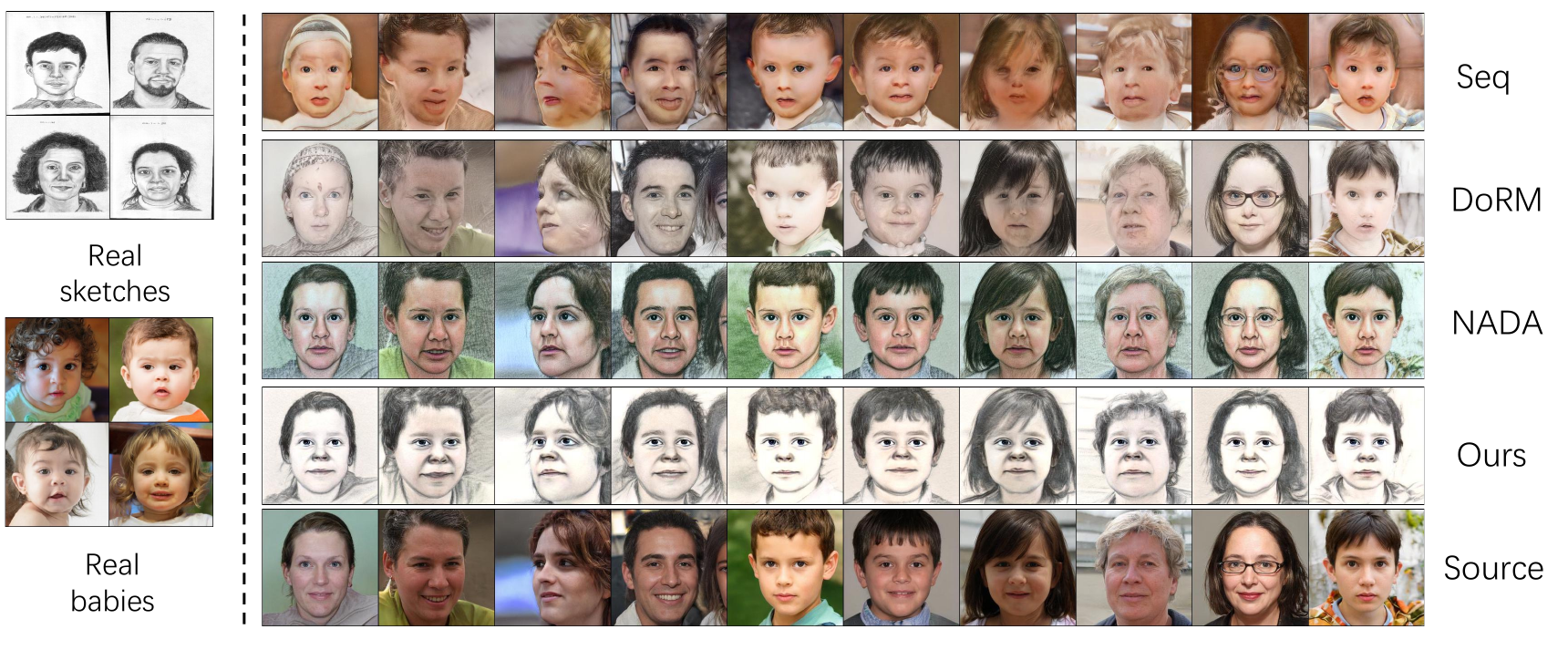}
    \end{minipage}
\label{fig:baby_sketch}
}
\\
\subfigure{
    \begin{minipage}[t]{1\textwidth}
    \centering
        \includegraphics[width=0.9\textwidth]{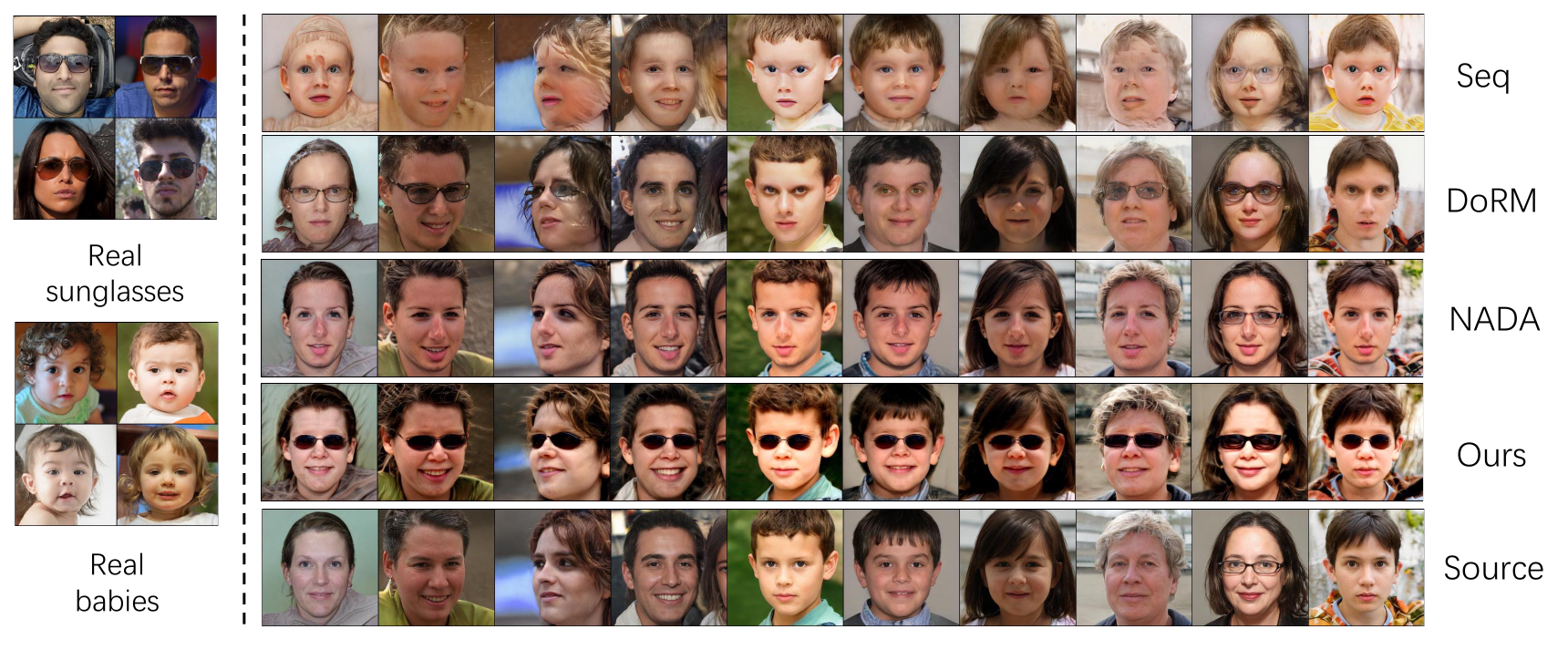} 
    \end{minipage}
    \label{fig:baby_sunglass}
}
 \vspace{-0.5cm}
\subfigure{
    \begin{minipage}[t]{1\textwidth}
    \centering
        \includegraphics[width=0.9\textwidth]{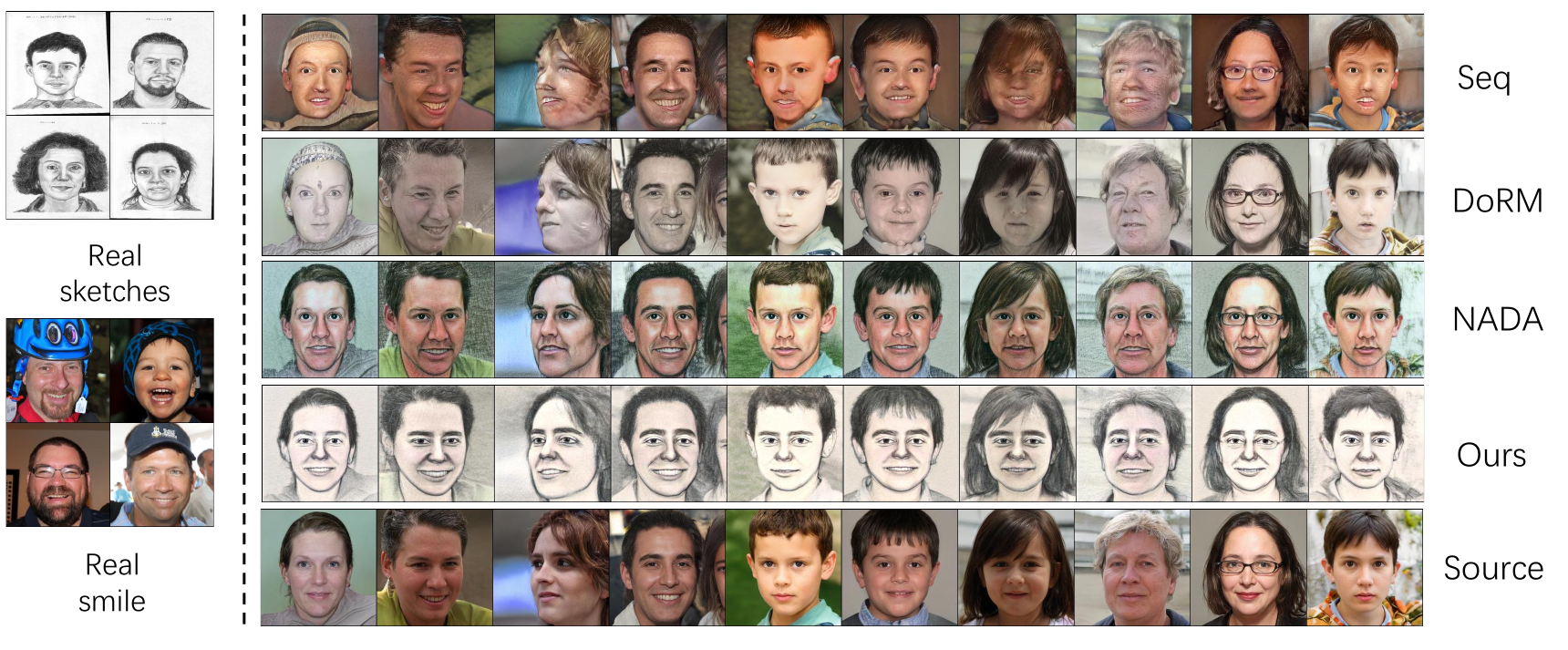} 
    \end{minipage}
    \label{fig:smile_sketch}
}
    \caption{Qualitative results on 10-shot HDA.}
    \label{fig:s-HDA}
    \vspace{-0.2cm}
\end{figure}

\begin{figure}[t]
    \centering
\subfigure{
    \begin{minipage}[t]{1\textwidth}
    \centering
    \includegraphics[width=0.9\textwidth]{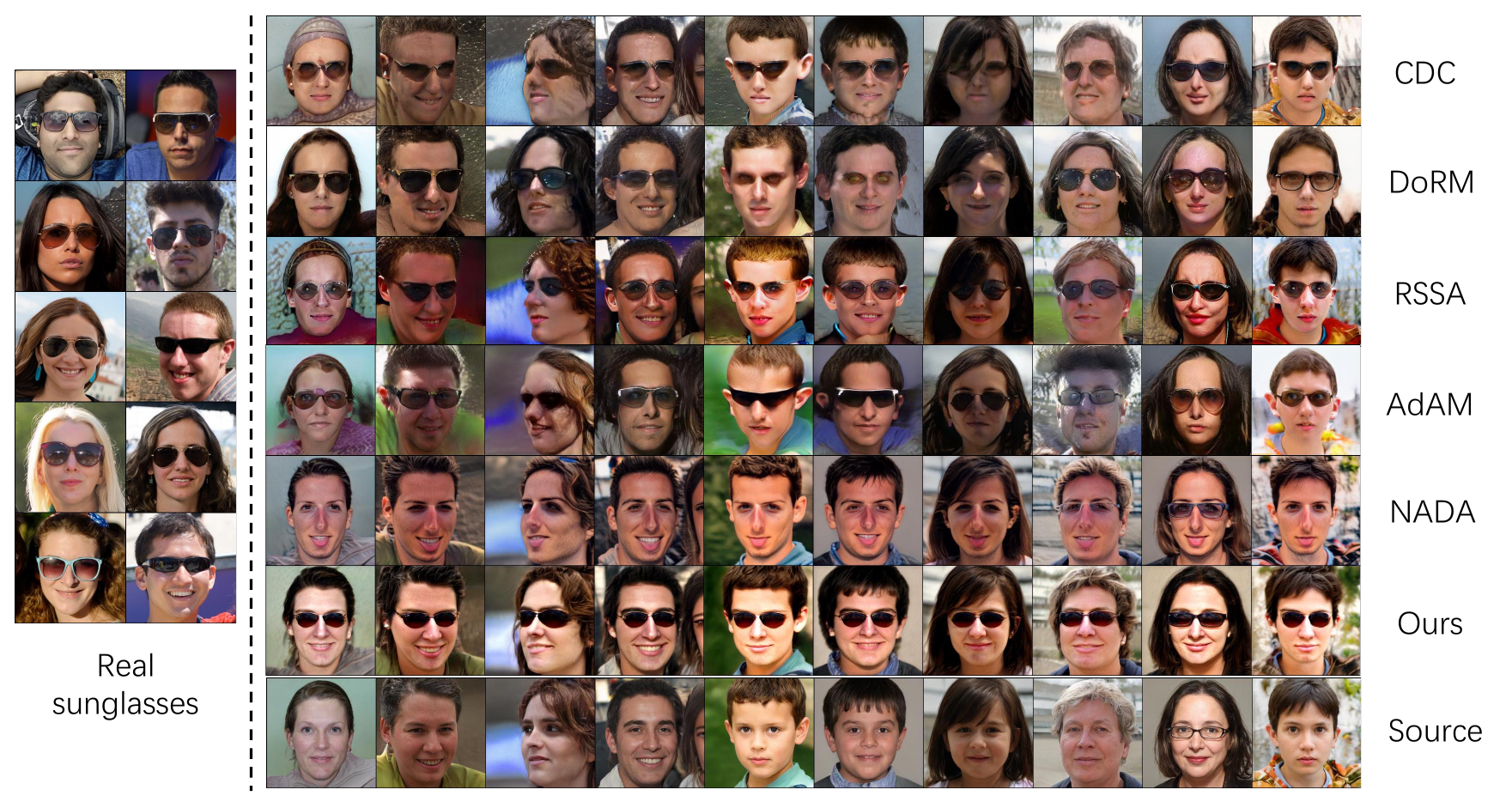}
    \end{minipage}
\label{fig:sunglass}
}
\\
 \vspace{-0.5cm}
\subfigure{
    \begin{minipage}[t]{1\textwidth}
    \centering
        \includegraphics[width=0.9\textwidth]{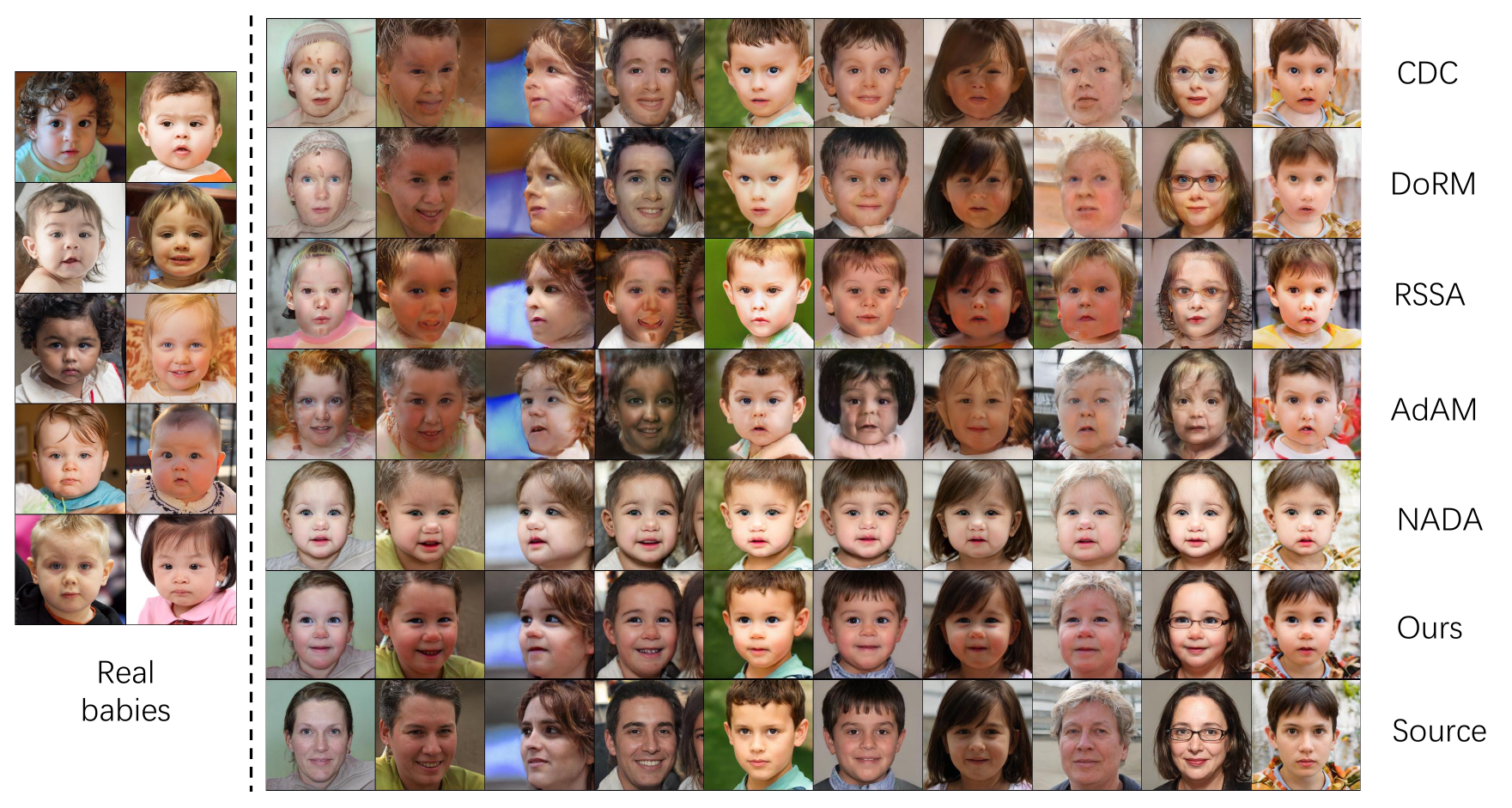} 
    \end{minipage}
    \label{fig:baby}
}
    \caption{Qualitative results on 10-shot DA of \textit{sunglasses} and \textit{baby}.}
    \label{fig:s-DA}
    \vspace{-0.2cm}
\end{figure}

\subsection{Intra-LPIPS}
Following CDC~\citep{ojha2021fig_cdc}, we utilize Intra-cluster pairwise LPIPS distance~\citep{zhang2018unreasonable} to evaluate the diversity of generated images for single domain adaptation. Specifically, we use 1000 generated images and 10 real images from the target domain to compute Intra-LPIPS. As shown in \cref{lpips}, our method achieves higher Intra-LPIPS distance than baseline methods, indicating more distinct images being generated.

\begin{table}
\small
  \centering
  \setlength{\tabcolsep}{4pt}
  \renewcommand\arraystretch{1.0}
\begin{tabular}{ccccccc}
\toprule
Intra-LPIPS ($\uparrow$)&\textit{smile}& \textit{sketch}&\textit{sunglasses}
\\
\midrule
NADA&
0.529 & 0.491& 0.535
\\
CDC&
0.616& 0.453& 0.562
\\
RSSA&
0.625& 0.480&	0.563
\\
AdAM&
0.615 & 0.495& 0.598	
 \\ 
Ours&	\textbf{0.642}& \textbf{0.506}&	\textbf{0.615}\\
 \bottomrule
\end{tabular}
\vspace{0.1cm}
\caption{Intra-cluster pairwise LPIPS distance on 10-shot DA.\label{lpips}%
}
\end{table}

\subsection{More Qualitative Results}
As the complement of qualitative results in the main paper, we conduct the experiments on more target domains. We present the results in \cref{fig:s-HDA} and \cref{fig:s-DA} for DA and HDA respectively. It can be observed that the generated images generated by our methods integrate the attributes, which significantly surpasses the baseline methods.

\subsection{Implement Details}
\label{sec:detail}
As depicted in \cref{hda-dist} and \cref{hda-direct}, we use pre-defined domain coefficient to modulate the attributes from multiple domains. For most experiments on two domains, we use $\alpha_i = 0.5$ except for \textit{baby-sunglasses} with $\alpha_{\textit{baby}} = 0.3$ and $\alpha_{\textit{sunglasses}} = 0.7$.
For the experiment of three domains, we set $\alpha_{\textit{baby}} = 0.7$,  $\alpha_{\textit{sketch}} = 0.4$, and $\alpha_{\textit{smile}} = 0.3$.

Following the setting of previous DA methods, we utilize the batch size 
 of 4 and a training session typically requires 300 iterations in roughly 3 minutes on a single NVIDIA TITAN GPU.
Besides, we set the balancing factor $\lambda$ as 1 in \cref{eq:da} and \cref{eq:hda}.

\subsection{Prompts for CLIP-Score}
To measure the semantic similarity with the target domain, we adopt CLIP-Score as the evaluation metric in \cref{metric}. As shown in \cref{prompt}, we present the prompts corresponding to different target domains. 
\begin{table}
\small
  \centering
  \setlength{\tabcolsep}{4pt}
  \renewcommand\arraystretch{1.0}
\begin{tabular}{c|c}
\toprule
Domain & Prompt\\
\midrule
\textit{baby} & ``face of a baby''
\\
\textit{sketch} & ``face of a person with the style of sketch''
\\
\textit{sunglasses} & ``face of a person with sunglasses''
\\
\textit{smile} & ``face of a person with smile''
\\
\textit{smile-sunglasses} & ``face of a person with smile and sunglasses''
\\
\textit{baby-smile} & ``face of a baby with smile''
\\
\textit{baby-sketch} & ``face of a baby with the style of sketch''
\\
 \bottomrule
\end{tabular}
\caption{Prompts of different domains for CLIP-Score.}
\label{prompt}
\end{table}

\begin{figure*}[t]
\centering
\includegraphics[width=0.95\linewidth]{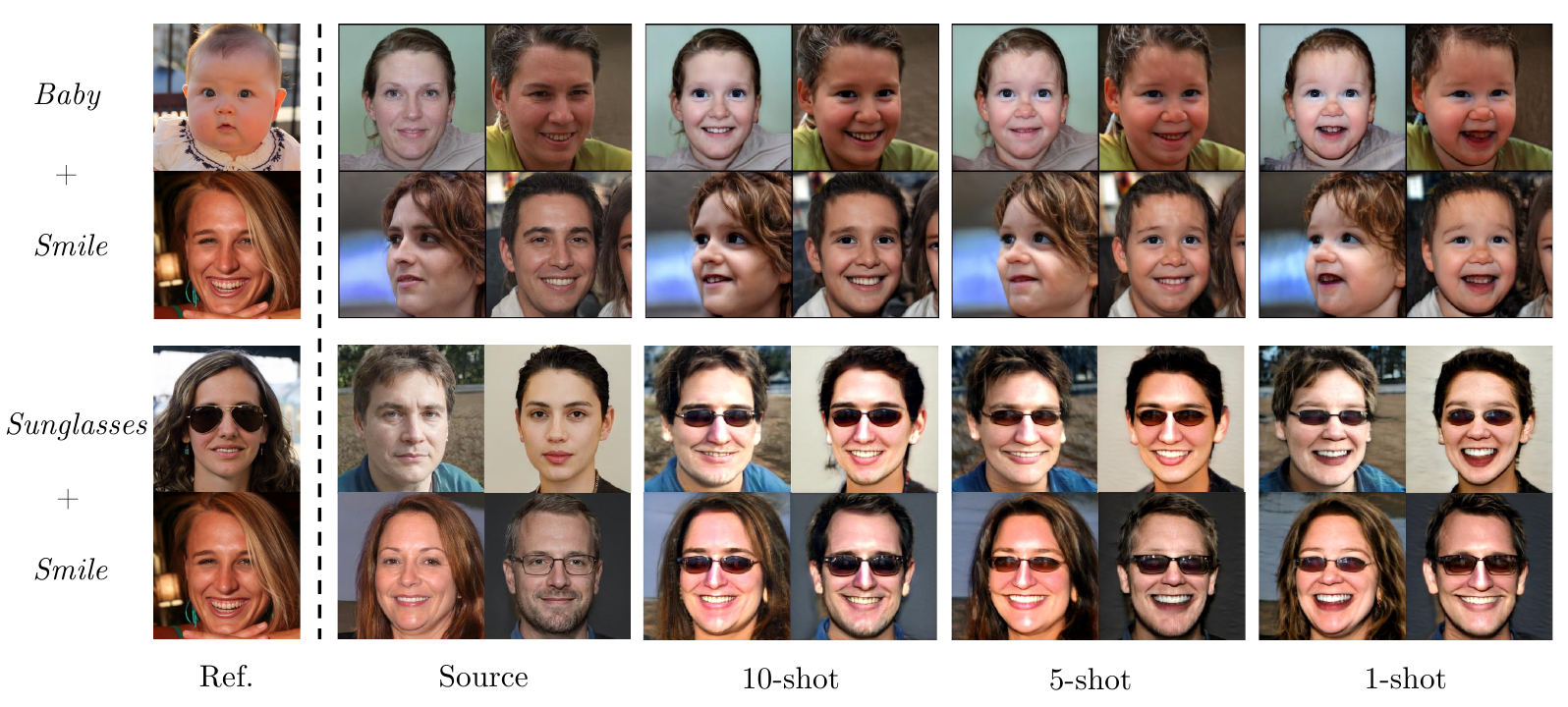}
\caption{\textcolor{black}{Impact of different shots for our method in HDA. Results of 5-shot are close to those of 10-shot. Compared with them, the results of 1-shot exhibit relatively lower cross-domain consistency.}}
\label{fig:shot}
\end{figure*}

\begin{figure*}[t]
\centering
\includegraphics[width=0.95\linewidth]{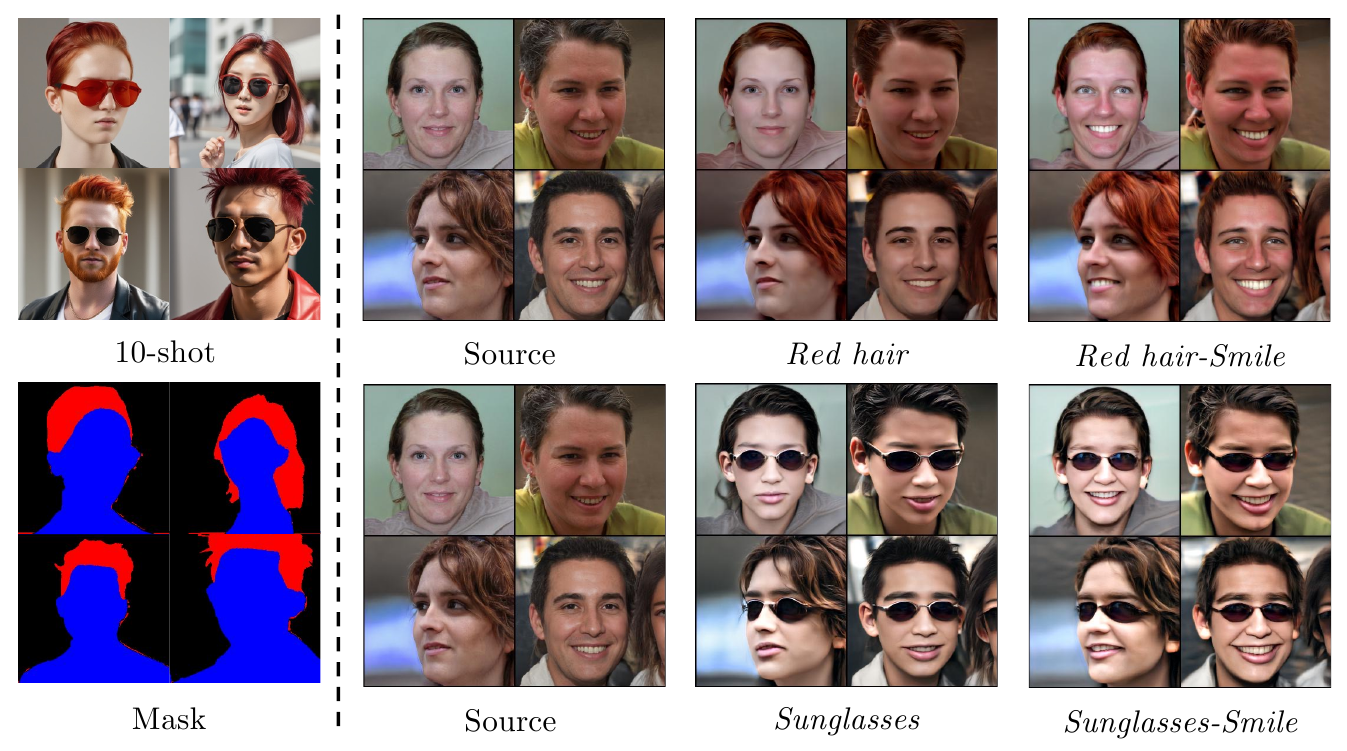}
\caption{\textcolor{black}{Conditional generation of our method in DA and HDA. We adapt the generator with masked images for \textit{red hair} and \textit{sunglasses} respectively. The red masks represent \textit{red hair}, while the blue masks represent faces wearing \textit{sunglasses}.}}
\label{fig:multi}
\end{figure*}

\textcolor{black}{
\subsection{Results using different numbers of shots}
We perform experiments on lower shots (5-shot and 1-shot) as suggested using various datasets. As shown in \cref{fig:shot}, the results of 5-shot are close to those of 10-shot, which integrates the attributes and maintains the consistency with source domain. Although multiple attributes have been learned, the results of 1-shot exhibit relatively lower cross-domain consistency compared with 10-shot and 5-shot.  This is because when there is only one reference image per domain, the subspace in our directional subspace loss degenerates into a single point (see \cref{fig:subspace}). Then distinct generated images corresponding to different noise tend to converge towards the same image in CLIP's embedding space, which comprises cross-domain consistency as depicted in \cref{sec:subspace}.
}

\textcolor{black}{
\subsection{Results of conditional generation}
As shown in \cref{fig:multi}, we conduct the experiments on conditional generation. Specifically, we collect 10-shot images with \textit{red hair} and \textit{sunglasses}. Then we use masks to separate these attributes and adapt the generator with masked images for \textit{red hair} and \textit{sunglasses} respectively. We can observe that the generated images possess the corresponding attribute for both single DA and hybrid DA. Simultaneously, these images also maintain consistency with source domain.
}

\begin{figure*}[t]
\centering
\includegraphics[width=0.95\linewidth]{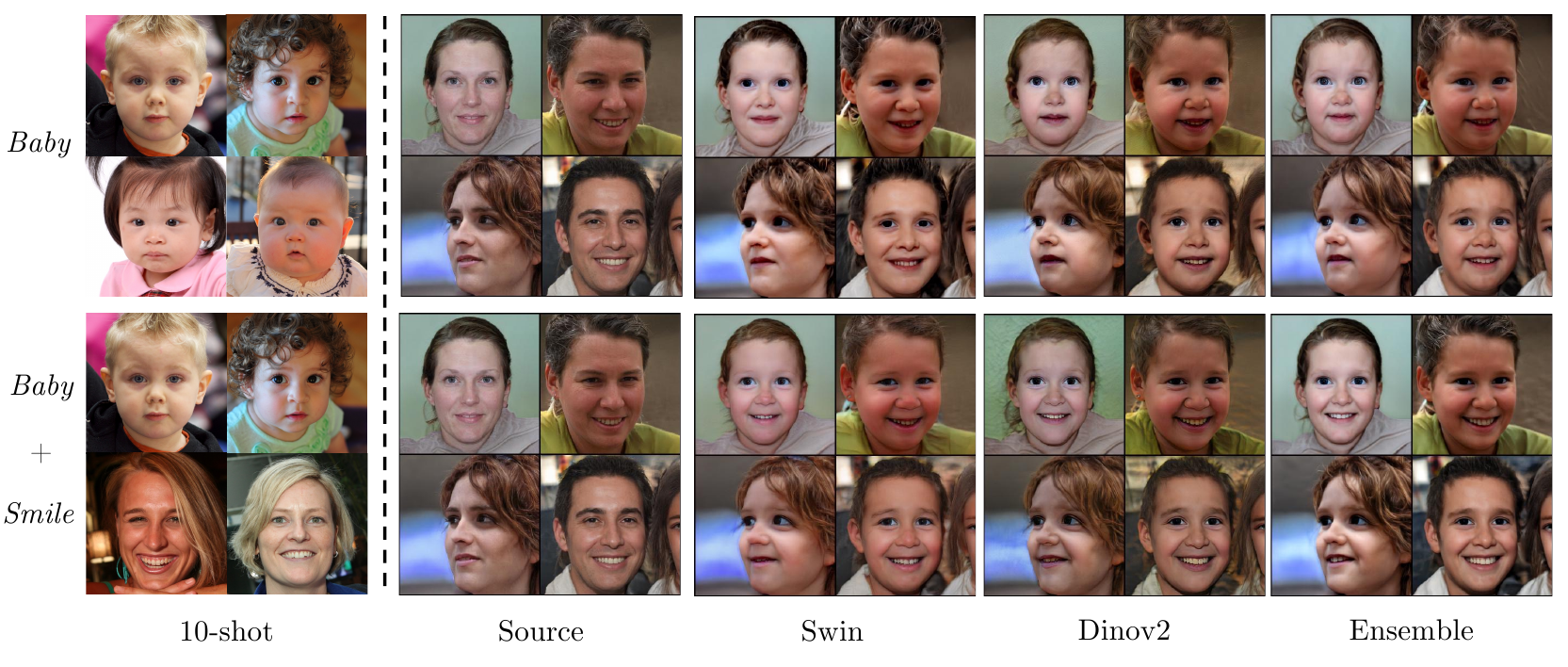}
\caption{\textcolor{black}{Effect of different pre-trained image encoders in \cref{fig:subspace}, \eg, Swin, Dinov2, and their ensemble.}}
\label{fig:encoder}
\vspace{-0.2cm}
\end{figure*}

\textcolor{black}{
\subsection{Effect of different pre-trained image encoders}
As shown in \cref{fig:encoder}, we conduct experiments on pre-trained Swin and Dinov2 to explore the impact of different image encoders on generated images. Our method is agnostic to different pre-trained image encoders. Although they exhibit slight stylistic differences, these are due to their different approaches to extract features into separated subspaces, as depicted in \cref{fig:tsne}. To converge to the exact realization of the target domain, our method employs the ensemble technique that exploits both Swin and Dinov2. As shown in the figure, the results closely resembles the attributes of the target domain while maintaining the best consistency with source domain.
}

\begin{figure*}[t]
\centering
\includegraphics[width=0.95\linewidth]{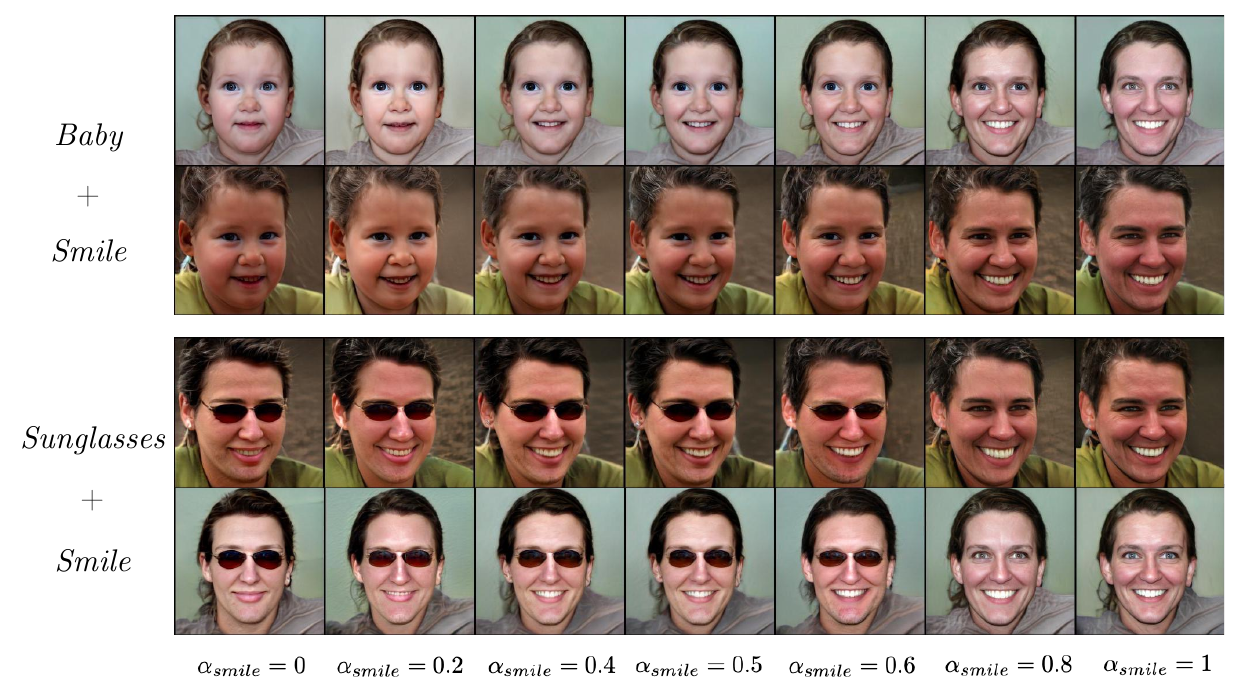}
\caption{\textcolor{black}{The study of simple traversal on $\alpha_i$ for hybrid domain adaptation. The attributes of generated images transit smoothly between two domains. Our method produces the similar attribute blending effect when $\alpha_i \in \{0.4, 0.5, 0.6\}$. Note that $\alpha_{baby}$ and $\alpha_{sunglasses}$ are $1 - \alpha_{smile}$.}}
\label{fig:traversal}
\end{figure*}

\textcolor{black}{
\subsection{Traversal for domain coefficient $\alpha$}
In our method for hybrid domain adaptation, this parameter controls the composition ratio of the attribute from each domain. As depicted in \cref{sec:detail}, we use $\alpha_i = 0.5$ for most experiments without the need for complex and intricate adjustments. To further explore the sensitivity, we conduct the study for simple traversal of $\alpha_i$. As shown in \cref{fig:traversal}, the attributes of generated images transit smoothly between domains. Our method produces the similar attribute blending effect when $\alpha_i \in \{0.4, 0.5, 0.6\}$.
}

\begin{figure*}[t]
\centering
\includegraphics[width=0.95\linewidth]{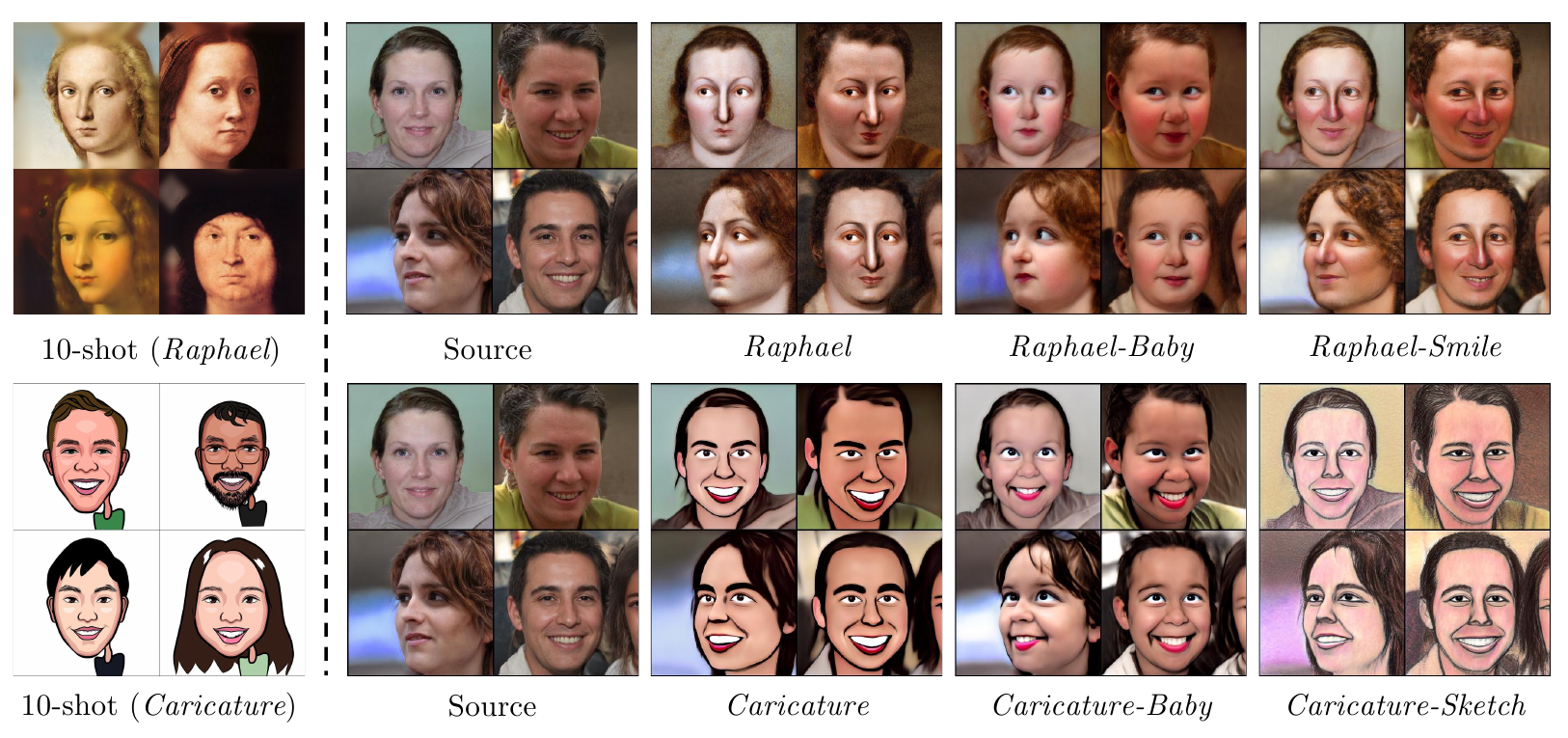}
\caption{\textcolor{black}{Single and hybrid domain adaptation on \textit{Raphael} and \textit{Caricature}.}}
\label{fig:cari}
\end{figure*}

\textcolor{black}{
\subsection{The results of more domains}
Consistent with prior work for few-shot generative domain adaptation (like CDC~\citep{ojha2021fig_cdc}, RSSA~\citep{xiao2022few} and DoRM~\citep{wu2023domain}), our experimental data encompasses both global style (\textit{sketch} and \textit{baby}) and local attributes (\textit{smile} and \textit{sunglasses}). The combination of these domains demonstrates the effectiveness of our method since they encompass all types of generative domain adaptation.
}

\textcolor{black}{
To provide more comprehensive evidence of validity, we conduct additional experiments on \textit{Raphael} and \textit{Caricature} for both single and hybrid domain adaptation. As shown in \cref{fig:cari}, the results integrate the characteristics from multiple target domains and maintains robust consistency with source domain, which further demonstrates the effectiveness of our method. 
}

\begin{figure*}[t]
\centering
\includegraphics[width=0.95\linewidth]{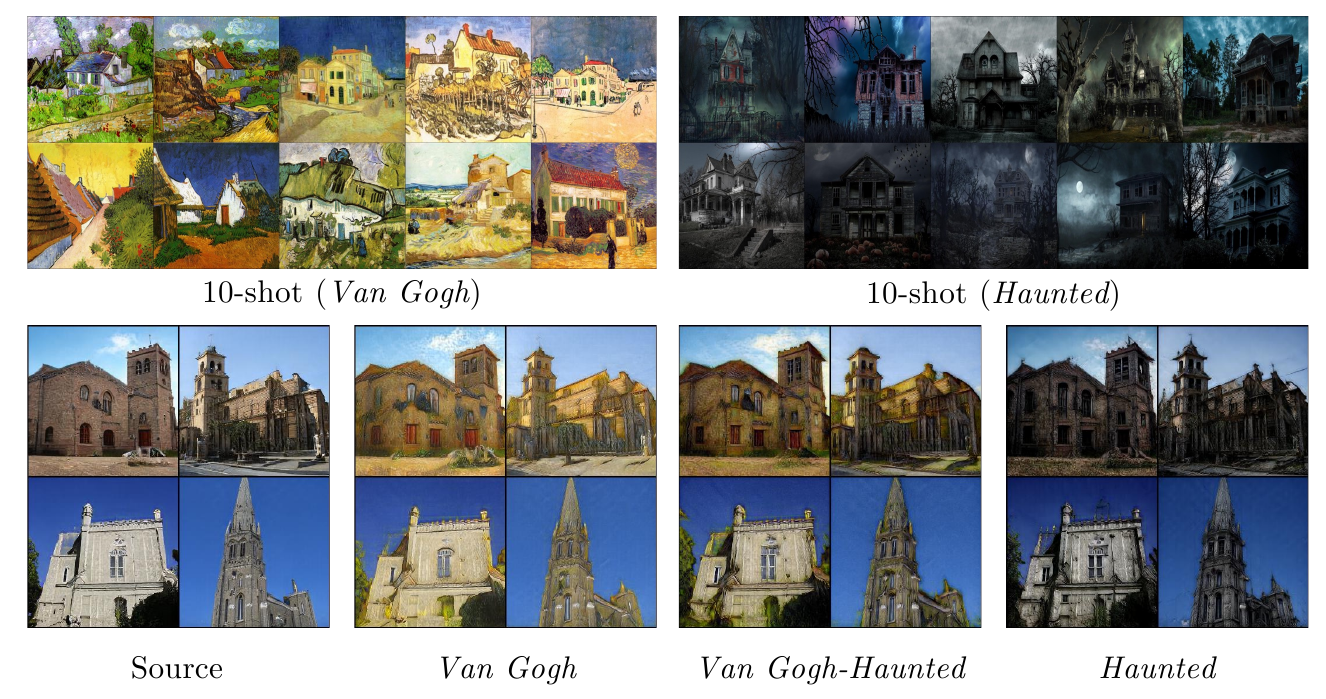}
\vspace{-0.3cm}
\caption{\textcolor{black}{Results of domain adaptation from \textit{LSUN Church} to \textit{Van Gogh's house paintings}, \textit{haunted houses}, and the combination of them.}}
\label{fig:church}
\vspace{-0.2cm}
\end{figure*}

\textcolor{black}{
\subsection{The generalizability to other domains}
To verify the generalizability of our method to other domains, we conduct experiments on church domain following prior work~\citep{ojha2021fig_cdc, xiao2022few, wu2023domain}, RSSA~\citep{xiao2022few} and DoRM~\citep{wu2023domain}). We adapt the pre-trained generator from \textit{LSUN Church} \citep{yu2015lsun} to \textit{Van Gogh's house paintings}, \textit{haunted houses}, and the combination of them. As shown in \cref{fig:church}, the results acquire the corresponding style and showcase the preservation of good cross-domain consistency. This aligns with the results observed in the face domain.
}

\begin{figure*}[t]
\centering
\includegraphics[width=0.95\linewidth]{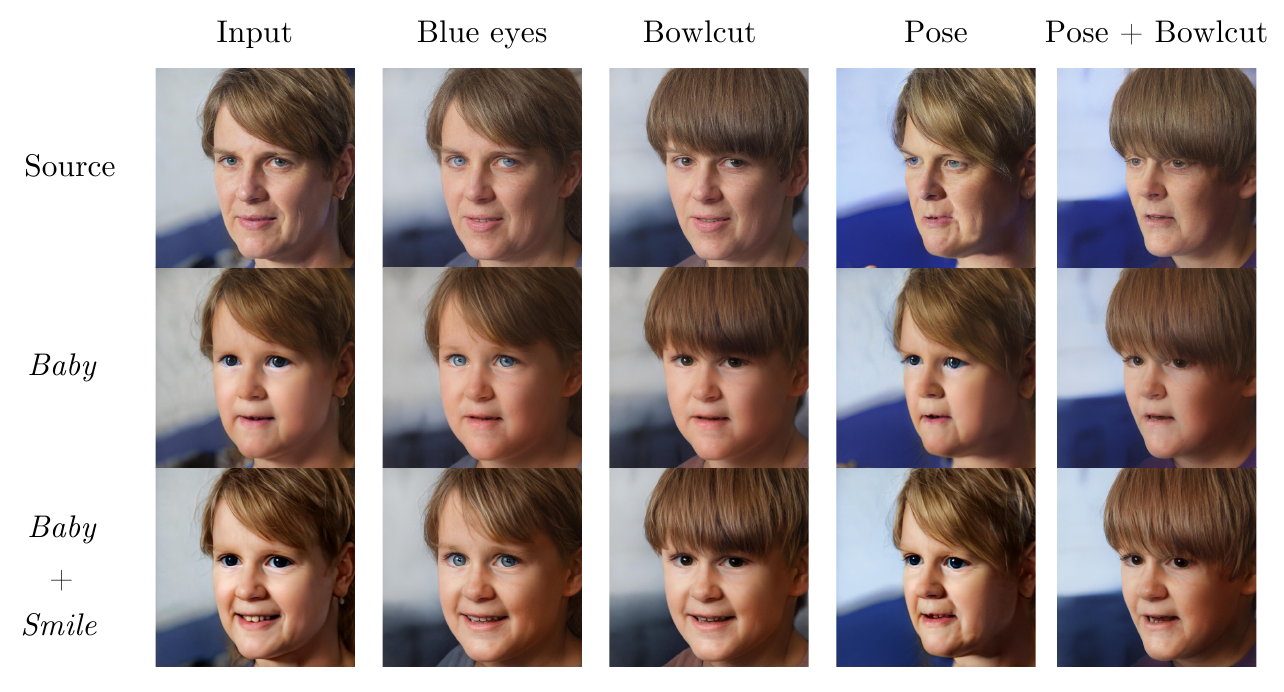}
\caption{\textcolor{black}{Editing results on the images generated by original and adapted model for both single and hybrid domain adaptation. The first line is the source images. The second and third lines are images of the adaptation.}}
\label{fig:edit}
\end{figure*}

\begin{figure*}[t]
\centering
\includegraphics[width=0.95\linewidth]{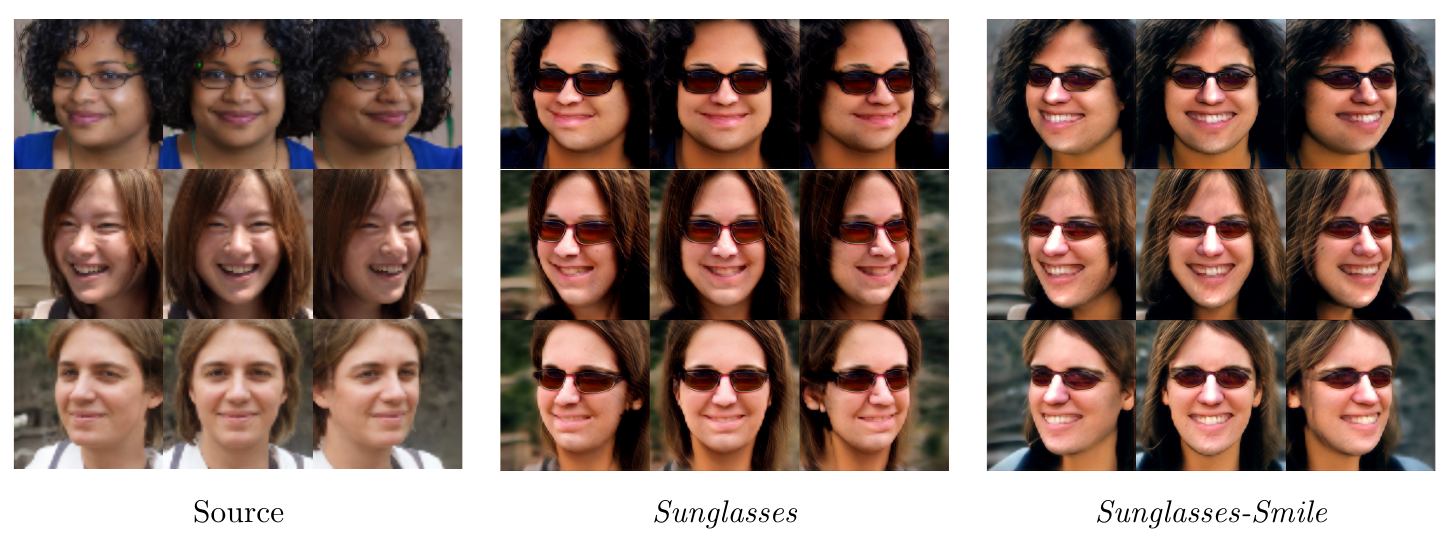}
\caption{\textcolor{black}{Single and hybrid domain adaptation in 3D GAN. We adapt pre-trained EG3D~\citep{chan2022efficient} with our method using 10-shot training images per domain.}}
\label{fig:3d}
\end{figure*}

\textcolor{black}{
\subsection{The editability before and after the domain adaptation}
We conduct the editing on the images generated by original and adapted model for both single and hybrid domain adaptation. As shown in the \cref{fig:edit}, the results indicate the adapted generator maintains similar editability like pose to the original generator. This verifies that the our method effectively preserves original generator's attributes. 
}

\textcolor{black}{
\subsection{Single and hybrid domain adaptation in 3D GAN.}
We conducted experiments using the popular 3D-aware image generation method, EG3D~\citep{chan2022efficient}. Specifically, we replace the discriminator as we did in \cref{fig:subspace} for both single and hybrid domain adaptation. As shown in \cref{fig:3d}, we adapt the pre-trained generator from FFHQ to \textit{sunglasses} and the hybrid of \textit{sunglasses} and \textit{smile} with 10-shot training images per domain. We can observe that the results effectively integrate the attributes and preserve the characters and poses of source domain.
}

\end{document}